\documentclass[10pt,journal,compsoc]{IEEEtran}
\usepackage{color}
\usepackage{graphicx}
\usepackage{booktabs}
\usepackage{enumitem}
\usepackage[export]{adjustbox}
\usepackage{multirow}
\usepackage{bm}

\usepackage{tabularray}

\usepackage{pifont}
\usepackage{makecell}

\usepackage{epsfig}
\usepackage{amsmath}
\usepackage{amssymb}
\usepackage{multirow}
\usepackage{mathrsfs}
\usepackage{footmisc}
\usepackage[colorlinks,linkcolor=red]{hyperref}
\usepackage{tabulary}
\usepackage[dvipsnames]{xcolor}
\usepackage{colortbl}
\usepackage{url}
\definecolor{mygray}{gray}{0.6}
\definecolor{mygray-bg}{gray}{0.93}
\def\eg{\emph{e.g}\onedot} 
\def\ie{\emph{i.e}\onedot}

\ifCLASSOPTIONcompsoc
  \usepackage[nocompress]{cite}
\else
  \usepackage{cite}
\fi
\usepackage{amsmath}
\usepackage{amssymb}

\usepackage{array}

\ifCLASSOPTIONcompsoc
 \usepackage[caption=false,font=footnotesize,labelfont=sf,textfont=sf]{subfig}
\else
 \usepackage[caption=false,font=footnotesize]{subfig}
\fi
\usepackage{url}

\makeatletter
\newcommand{\thickhline}{%
	\noalign {\ifnum 0=`}\fi \hrule height 1pt
	\futurelet \reserved@a \@xhline
}
\makeatother

\newcommand{\app}{\raise.17ex\hbox{$\scriptstyle\sim$}}

\newcolumntype{x}[1]{>{\centering\arraybackslash}p{#1pt}}
\newlength\savewidth

\def\eg{\emph{e.g}.} 
\def\ie{\emph{i.e}.} 
\def\etc{\emph{etc}.} 
 
\def\etal{\emph{et al}.}

\begin{document}
%
\title{Towards Vehicle-to-everything Autonomous Driving: A Survey on Collaborative Perception}
%
%
%
%

\author{Si~Liu, 
        Chen~Gao, 
        Yuan~Chen, 
        Xingyu~Peng, 
        Xianghao~Kong, 
        Kun~Wang, \\
        Runsheng~Xu, 
        Wentao~Jiang, 
        Hao~Xiang, 
        Jiaqi~Ma, 
        Miao~Wang
\IEEEcompsocitemizethanks{\IEEEcompsocthanksitem Si Liu, Chen Gao, Yuan Chen, Xingyu Peng, Xianghao Kong, Kun Wang, and Wentao Jiang are with Beihang University. E-mail: \{liusi, gaochen, chenyuan1, pengxyai, refkxh, jiangwentao\}@buaa.edu.cn. Corresponding author: Si Liu. 
\IEEEcompsocthanksitem Runsheng Xu, Hao Xiang, and Jiaqi Ma are with UCLA Mobility Lab. E-mail: rxx3386@ucla.edu, haxiang@g.ucla.edu, and jiaqima@ucla.edu.
\IEEEcompsocthanksitem Kun Wang and Miao Wang are with Beijing Baidu Apollo Mobility Technology Co., Ltd., Beijing, China.
}
}


%
%

\markboth{Journal of \LaTeX\ Class Files,~Vol.~14, No.~8, August~2021}%
{Shell \MakeLowercase{\textit{et al.}}: A Sample Article Using IEEEtran.cls for IEEE Journals}
%



\IEEEtitleabstractindextext{%
\begin{abstract}
   Vehicle-to-everything (V2X) autonomous driving opens up a promising direction for developing a new generation of intelligent transportation systems. Collaborative perception (CP) as an essential component to achieve V2X can overcome the inherent limitations of individual perception, including occlusion and long-range perception.  
   In this survey, we provide a comprehensive review of CP methods for V2X scenarios, bringing a profound and in-depth understanding to the community. 
   Specifically, we first introduce the architecture and workflow of typical V2X systems in practice, which affords a broader perspective to understand the entire V2X system and the role of CP within it. Then, we thoroughly summarize and analyze existing V2X perception datasets and CP methods. Particularly, we introduce numerous CP methods from various crucial perspectives, including collaboration stages, roadside sensors placement, latency compensation, performance-bandwidth trade-off, attack/defense, pose alignment, \etc~Moreover, we conduct extensive experimental analyses to compare and examine current CP methods, revealing some essential and unexplored insights.
   Specifically, we analyze the performance changes of different methods under different transmission bandwidths, providing a deep insight into the performance-bandwidth trade-off issue. Also, we examine methods under different LiDAR ranges.
   To study the model robustness, we further investigate the effects of various simulated real-world noises on the performance of different CP methods, covering communication latency, lossy communication, localization errors, and mixed noises. 
   In addition, we look into the sim-to-real generalization ability of existing CP methods. 
   At last, we thoroughly discuss issues and challenges, highlighting promising directions for future efforts. Our codes for experimental analysis will be public at \url{https://github.com/memberRE/Collaborative-Perception}.
\end{abstract}

\begin{IEEEkeywords}
  Collaborative Perception, Vehicle-to-Everything, Deep Learning, Survey.
\end{IEEEkeywords}}

\maketitle

\IEEEdisplaynontitleabstractindextext

%
\IEEEpeerreviewmaketitle

\IEEEraisesectionheading{\section{Introduction}\label{sec:introduction}}


%
%
%
%

\IEEEPARstart{A}{utonomous} driving technology~\cite{huang2019apolloscape,singh2022road} has experienced remarkable progress, which has the potential to revolutionize the transportation industry as it can effectively increase driving safety/efficiency, and reduce traffic incidents.
Particularly, perception is one of the most critical abilities to achieve robust autonomous driving, enabling vehicles to accurately perceive and understand the surroundings, and drive accordingly. 
With the advancement of deep learning, machine perception methods demonstrate impressive abilities in various tasks, including 2D/3D object detection~\cite{zou2023object,wang2021salient,guo2020deep,zhu2021cylindrical}, segmentation~\cite{liao2022kitti,minaee2021image,zhou2022survey}, tracking~\cite{smeulders2013visual,javed2022visual}, \etc~
Although these advancements significantly improve the perception capability of autonomous vehicles, the perception paradigm still lies in single-vehicle/individual perception, \ie, vehicles rely on their onboard sensors and computing devices to complete perception tasks.

\begin{figure}[t]
    \begin{center}
    \includegraphics[width=0.95\linewidth]{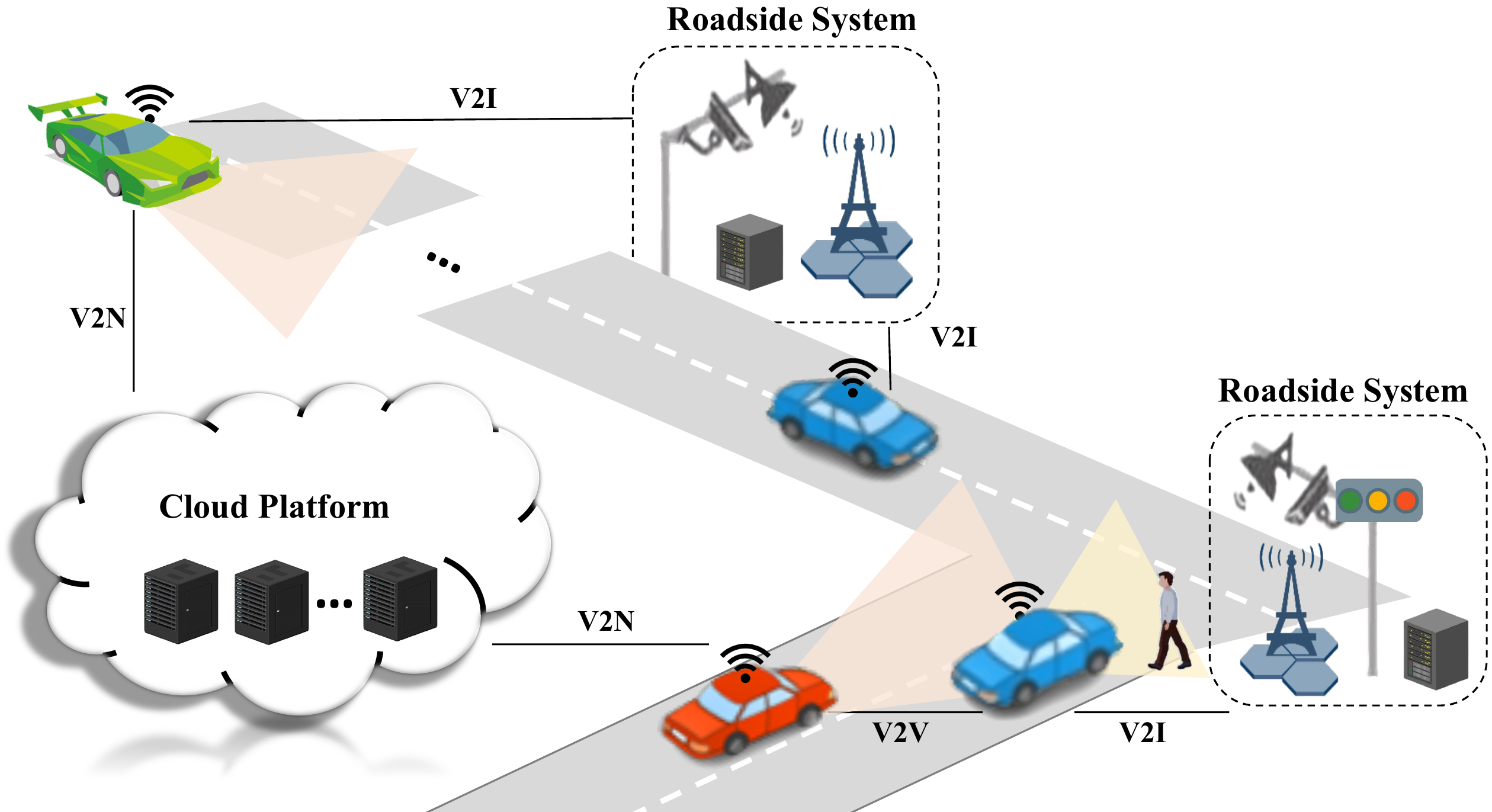}
    \end{center}
    \vspace{-5mm}
    \caption{A diagram illustrating V2X scenarios. The red car faces the occlusion issue, and the green car faces the long-range perception issue. By obtaining extra perceptual information from other vehicles (V2V) or infrastructure (V2I), these vehicles can achieve a holistic perception of their surroundings, improving traffic safety.}
    \vspace{-4mm}
    \label{fig:fig1}
\end{figure}
In fact, single-vehicle perception is hard to meet the demand of high-level autonomous driving for precise perception, facing inevitable limitations and challenges. For example, (\romannumeral1) vehicles are generally constrained by cost and space limitations. Thus vehicles typically are equipped with low-precision sensors and low-power computing devices, which limits the perception ability of an autonomous vehicle. (\romannumeral2)  As shown in Fig.~\ref{fig:fig1} (red car), due to the obstruction of other vehicles or obstacles, single-vehicle perception can only obtain limited sight-of-view. Thus it is difficult to achieve a holistic perception of the surroundings, leading to potential traffic accidents. (\romannumeral3) As shown in Fig.~\ref{fig:fig1} (green car), vehicles need to perceive objects or conditions at long distances, especially in high-speed driving scenarios. However, long-range objects typically exhibit sparsity in sensor data, \eg, occupying only a few pixels in camera images or a small number of points in LiDAR point clouds, making it prone to erroneous perception results.

\begin{figure*}[t]
    \begin{center}
    \includegraphics[width=1\linewidth]{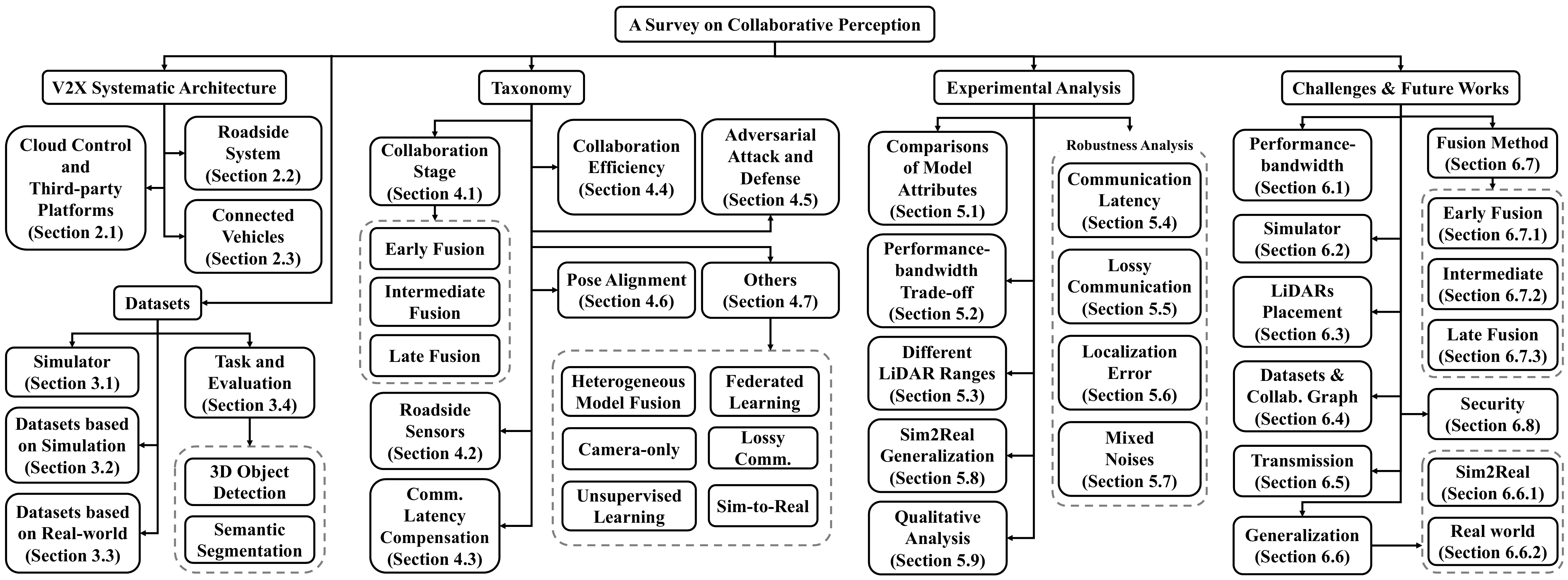}
    \end{center}
    \vspace{-4mm}
    \caption{Overview of this survey.}
    \vspace{-4mm}
    \label{fig:paper-structure}
\end{figure*}
In recent years, vehicle-to-everything (V2X) autonomous driving has attracted a surge of interest within both academic and industrial communities. V2X refers to the vehicle can share complementary information with other traffic elements by communication technology to achieve more accurate and safety self-driving. Concretely, according to the different subjects and objects of information transmission, V2X can be divided into vehicle-to-vehicle (V2V), vehicle-to-infrastructure (V2I), vehicle-to-pedestrian (V2P), and vehicle-to-network (V2N) as shown in Fig.~\ref{fig:fig1}. 
Moreover, collaborative perception (CP) is the essential technology to achieve V2X, enabling vehicles to exchange perceptual information with other traffic elements and obtain a holistic perception of surroundings. Thus CP can essentially overcome the bottleneck issue faced by individual perception. For example, vehicles can leverage the powerful computing resources on the cloud platform by V2I/V2N to efficiently execute large-scale and regularly updated perception models. Particularly, through obtaining extra perceptual information from other vehicles and infrastructures, as shown in Fig.~\ref{fig:fig1}, vehicles can overcome the occlusion and long-range perception issues faced by individual perception and achieve beyond-line-of-sight perception capability.

Numerous papers are produced to drive the rapid development of this field. 
Large-scale datasets are essential for training deep-learning models, and most perception datasets are annotated for individual perception specifically. Thus pioneer researchers put a significant amount of effort into building CP datasets for V2X scenarios. Unlike other perception tasks that can easily collect data from the internet, obtaining real-world V2X perception data and building benchmarks~\cite{yu2022dair,xu2023v2v4real} is far more difficult due to excessive costs. Thus, many works~\cite{li2022v2x,xu2022opv2v} also collect and annotate simulated data based on various traffic simulators~\cite{dosovitskiy2017carla,xu2021opencda}.
Moreover, V2X collaborative perception is a multi-agent system that involves information transmission and fusion among multiple agents, which presents unique challenges compared to individual perception. 
From the perspective of transmission efficiency, some works~\cite{huwhere2comm,fpvrcnn} are devoted to improving performance while reducing transmission bandwidth. From the perspective of information fusion stages, most works~\cite{wang2020v2vnet,xu2022v2x,huwhere2comm,xiang2023hm} choose to fuse intermediate features, and some works fuse raw data~\cite{chen2019cooper} or final results~\cite{shi2022vips}. From the perspective of model robustness, some works attempt to simulate and address communication time delays~\cite{qiu2021autocast}, losses~\cite{xu2022v2x},  attacks~\cite{li2023among}, localization errors~\cite{fpvrcnn}, and other issues that may arise in real-world environments. 
Compared to related surveys~\cite{ren2022collaborative,bai2022survey,han2023collaborative}, this paper provides a more comprehensive summary and deeper analysis in the collaborative perception field. In addition to the thorough taxonomy of various datasets and methods, we present the practical V2X system architecture and conduct extensive experiments from various aspects to analyze crucial yet under-explored issues, bringing insightful discussions and promising future directions. The organization of this paper is shown in Fig.~\ref{fig:paper-structure}.

\begin{figure*}[t]
    \begin{center}
    \includegraphics[width=0.95\linewidth]{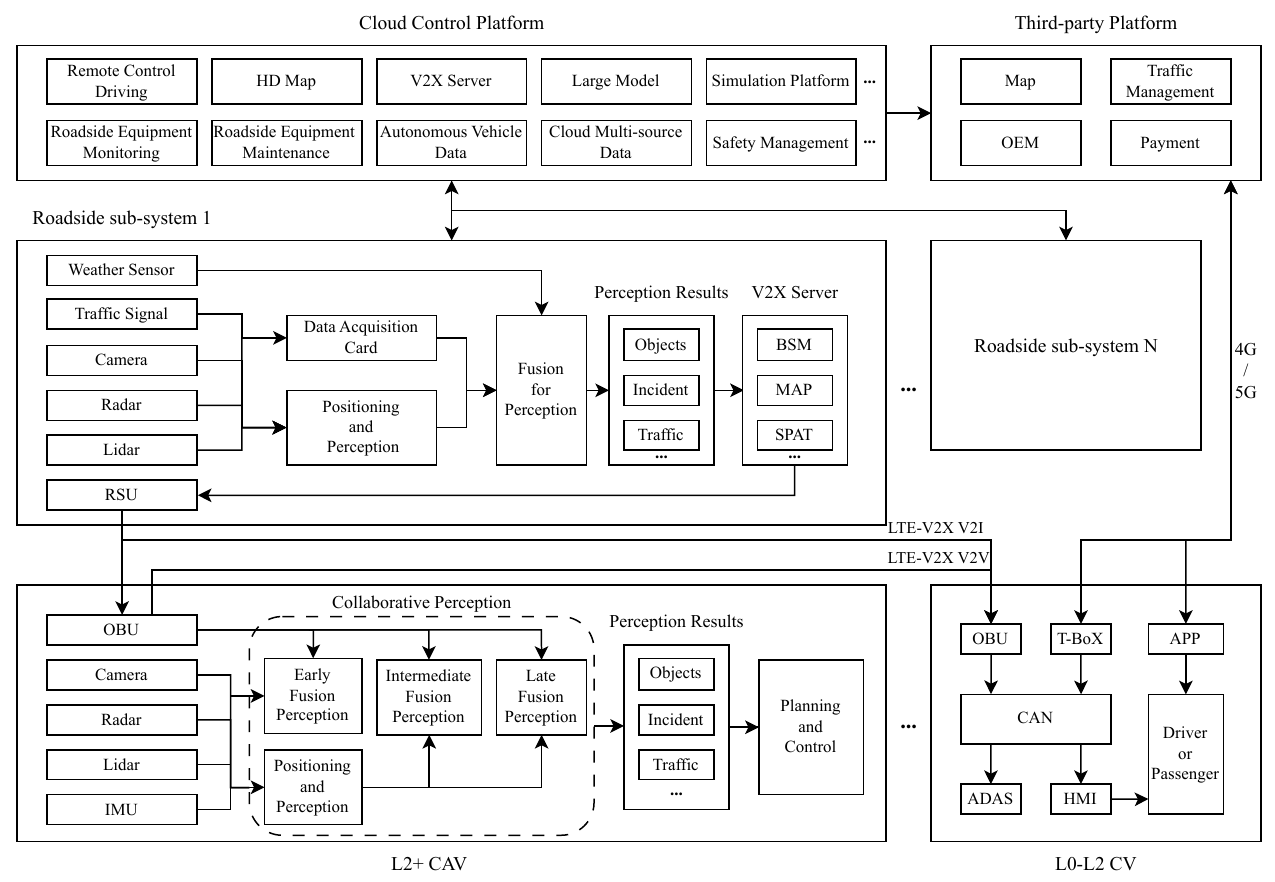}
    \end{center}
    \vspace{-6mm}
    \caption{The illustration of a typical V2X system architecture in practical applications.}
    \vspace{-4mm}
    \label{fig:sys-flow}
\end{figure*}

To sum up, the main contributions of this paper can be summarized as follows:
\begin{enumerate}
    \item We present the architecture and workflow of a typical V2X system in practical applications, aiming to provide researchers with a clearer perspective on the entire V2X system and the role of CP methods within it.
    \item We provide a comprehensive literature review on V2X datasets and methods, summarizing datasets and categorizing methods from various perspectives. The review and taxonomy afford an in-depth understanding of the critical factors of CP methods in V2X scenarios.
    \item We test and compared state-of-the-art models from various aspects such as model size and efficiency, providing a broad view of existing methods.
    \item We conduct extensive experiments to thoroughly investigate the robustness of current CP methods against widespread interference in the real world by simulating various types of noises such as time delay, lossy communication, localization errors, and mixed noises. 
    \item We conduct performance-bandwidth trade-off experiments and give an intensive comparison of various CP methods by adjusting the communication volume. We also provide cross-domain generalization studies to examine the crucial sim-to-real ability.  
    \item The comprehensive studies uncover the pros/cons of current methods and can bring new observations and insights to the community. We present a thorough discussion about open issues and potential directions based on our investigation and experimental analysis, to facilitate future academic and industry research.
\end{enumerate}

\section{V2X Systematic architecture}
\label{sec:sys}

In this section, we introduce a typical V2X system from an architecture perspective as shown in Fig.~\ref{fig:sys-flow}, aiming to provide the workflow in practice and a clearer understanding about the role of cooperative perception (CP) in the entire V2X system. The V2X system typically contains four parts, \ie, cloud control platform, third-party platform, roadside system, and connected vehicles (CVs). Note that CVs contain connected autonomous vehicles (CAVs) that have more advanced autonomous driving capabilities.

\subsection{Cloud Control and Third-party Platforms}
The cloud control platform has a powerful computing capability that can provide centralized control and management for the entire V2X network. 
Specifically, the remote control driving function enables well-trained operators to remotely drive CVs, which can provide emergency takeover and L4 autonomous driving capability. 
The cloud platform maintains High-Definition Map (HD Map) that provides CVs with highly accurate and detailed geographic information, including road geometry, terrain elevation, lane markings, traffic signals, road signs, obstacles, \etc~HD map is crucial for ensuring safe and high-level autonomous driving and updated regularly to stay up-to-date road network.
The cloud platform constantly collects and stores multi-source data from the roadside system and CVs for subsequent processing and applications. For example, the platform monitors signals from the roadside system and performs status analysis and safety management for maintaining the normal operation of roadside equipment. Besides, with sufficient storage and computing resources in the cloud platform, large models can be developed to support various tasks such as perception, prediction, planning, and simulation.

The third-party platform is also important in the V2X ecosystem, which provides value-added services (\eg, payment) to users, and facilitates data exchange between stakeholders. 
The cloud platform can provide multi-source data to the suppliers in the third-party platform. For example, the original equipment manufacturer (OEM) obtains roadside equipment data from the cloud platform for equipment upgrade and maintenance. The traffic data like traffic lights or incidents can be transmitted to the map supplier.
Through the internet, map suppliers can provide real-time map and navigation services to the in-car or mobile-phone APP for drivers.
Also, with large-scale traffic data, government agencies can improve the level of traffic management.

\subsection{Roadside System}
The roadside system typically has communication and perception capabilities and is a critical component for achieving V2X AD, enabling CVs to communicate with each other~(V2V) and surrounding infrastructures~(V2I).
The entire roadside system typically consists of multiple sub-systems, with each sub-system responsible for a specific area, much like a cellular network. 

The computational power of the roadside system lies between the cloud platform and the vehicle. With more physical space than the vehicle, the roadside system can accommodate more high-precision and high-power sensors along the roadways such as weather sensors, cameras, radar, lidar, \etc~Deployed algorithms (\eg, detection, segmentation, and tracking) can perform positioning and perception of specific objects on the road based on signals from the visual sensors. Besides, the data acquisition card can collect data from various types of traffic signals (\eg, traffic lights and traffic signs) and then convert the analog signals from these input channels into digital data for subsequent processing and analysis.
The fusion module adjusts and produces the perception results by combining information from the acquisition card and visual perception. Typically, the perception results contain real-time objects' location and category, incident recognition, traffic flow, \etc~

The V2X server converts the perception results into communication messages and transmits them to CVs/cloud through the dedicated bandwidth frequency of the wireless communication, \ie, $5905$-$5925$ MHz for DSRC-based systems and $5855$-$5925$ MHz for C-V2X-based systems. For example, Basic Safety Message (BSM) is a standardized message format used to transmit basic safety information such as a vehicle's position, speed, and direction. Map Data Attribute Profile (MAP) is a data attribute protocol used for transmitting high-precision map data. Signal Phase and Timing (SPAT) is a standardized message format used to transmit signal phases (green, yellow, and red lights) and timing of signal changes to vehicles.
The Roadside Unit (RSU) is the communication hub, collecting and relaying messages between CVs, the cloud, and other RSUs.

\subsection{Connected Vehicles (CVs)}
CVs are vehicles that are equipped with advanced communication technologies. On-Board Unit (OBU) is the communication device installed in CVs and adopts wireless communication technologies to exchange messages with other CVs, the roadside system, and the cloud platform. The L2-L4 connected autonomous vehicle (CAV) has more advanced autonomous driving capabilities than the L0-L2 CV, including perception and planning.

CVs are constrained by cost and space limitations, resulting in carrying low-precision sensors and low-power computing devices compared to the cloud platform and the roadside system, including camera, radar, lidar, and inertial measurement unit (IMU). The IMU consists of accelerometers, gyroscopes, and magnetometers, that can measure a vehicle's acceleration, orientation, and angular velocity. 

Moreover, the key difference between CAVs and normal AVs lies in the ability of CAVs to implement collaborative perception, \ie, CAVs combine the complementary visual information from the roadside system and the cloud platform to improve the perception ability. Specifically, divided by the fusion stage, there are three types of collaborative methods: early, intermediate, and late fusion. 
(\romannumeral1) In early fusion perception, the raw sensor data (\eg, RGB images and lidar point cloud) of the roadside system and other CVs are transmitted and received by RSU or OBU. Then the raw data are combined before any individual processing or feature extraction, and the perception models take the combined raw data as input to produce results. Thus early fusion provides better robustness but consumes larger bandwidth to transmit large amounts of raw data. 
(\romannumeral2) In contrast to early fusion, the intermediate fusion method combines information at the feature level. Other CVs and the roadside system need to transmit processed and extracted features, which fuse with the intermediate features of the ego vehicle's own perception model. This approach can reduce data transmission volume, leading to a more efficient and flexible collaborative perception.
(\romannumeral3) In late fusion perception, the independent perception results of ego vehicle, other CVs, and roadside system are transmitted and fused together to generate final results. For example, when performing object detection, the results of object bounding boxes are transmitted and can be fused via Non-Maximum Suppression (NMS). The late fusion paradigm has the lowest transmission bandwidth requirement and is easier to implement in practical systems than early and intermediate fusion. Thus late fusion is currently the predominant solution in the industry. 
However, late fusion suffers from poor accuracy and robustness, as it relies on independent results that may contain missed and false detection, leading to error accumulation.
The perception results contain object localization, incident, traffic flow recognition, \etc~Then the planning and control module produces driving decisions and driving actions based on the various perception results.

\begin{table*}[t]
   \caption{A summary of existing datasets for collaborative perception in V2X scenarios. Most datasets are built upon various traffic simulators, and some datasets collect data from the real world. See \S\ref{sec:datasets} for more details.}
   \vspace{-6mm}
   \begin{center}
      \resizebox{\linewidth}{!}{
         \renewcommand\arraystretch{1.3}
    \footnotesize
    \setlength{\tabcolsep}{3.5pt}{
        \begin{tabular}{c|cccccccccccc}
		\hline
            \multirow{2}{*}{Dataset} & \multirow{2}{*}{Publication} & \multirow{2}{*}{Source} & \multirow{2}{*}{Scenario} & \multicolumn{3}{c}{Sensor} & \multicolumn{3}{c}{Tasks} & \multirow{2}{*}{Frames} & \multirow{2}{*}{Viewpoints} & \multirow{2}{*}{Link} \\
            & & & & RGB&Depth&LiDAR&Detection	& Tracking&	Segmentation\\
		\hline
			VANETs~\cite{maalej2017vanets} & GLOBECOM 2017 & KITTI~\cite{geiger2012we} & V2V & \checkmark &  & \checkmark & \checkmark  &  &  & - & - & -\\
            MFSL~\cite{xiao2018multimedia} & ICMEW 2018 & KITTI~\cite{geiger2012we} & V2V & \checkmark &  &  &  &  & \checkmark  &  -& - & - \\
            T\&J~\cite{chen2019cooper} & ICDCS 2019 & Real-World & V2V &  &  & \checkmark & \checkmark &  & & 100 & 2 & \href{https://github.com/Aug583/F-COOPER}{Link}\\
            V2V-Sim~\cite{wang2020v2vnet} & ECCV 2020 & LiDARsim~\cite{manivasagam2020lidarsim} & V2V &  &  & \checkmark & \checkmark & \checkmark &  & 51,200 & - & -\\
            CoopInf~\cite{arnold2020cooperative} & TITS 2020 & CARLA~\cite{dosovitskiy2017carla} & V2I & \checkmark & \checkmark &  & \checkmark & \checkmark &  & 10,000 & 6,8&\href{https://github.com/eduardohenriquearnold/coop-3dod-infra}{Link}\\
            WIBAM~\cite{howe122021weakly} & BMVC 2021 & Real-World & V2I & \checkmark &  &  & \checkmark & \checkmark &  & 33,092 & 2-4 &\href{https://github.com/MatthewHowe/WIBAM}{Link}\\
            CODD~\cite{arnold2021fast} & RAL 2021 & CARLA~\cite{dosovitskiy2017carla} & V2V &  &  & \checkmark & \checkmark & \checkmark &  &  8,783 & 10 (avg.)&\href{https://github.com/eduardohenriquearnold/CODD}{Link}\\
            V2X-Sim~\cite{li2022v2x} & RAL 2022 & CARLA~\cite{dosovitskiy2017carla} \& SUMO~\cite{krajzewicz2012recent} & V2V,V2I & \checkmark & \checkmark & \checkmark & \checkmark & \checkmark & \checkmark & 10,000 & 2-5&\href{https://ai4ce.github.io/V2X-Sim}{Link}\\
            COMAP~\cite{yuan2021comap} & RAL 2022 & CARLA~\cite{dosovitskiy2017carla} \& SUMO~\cite{krajzewicz2012recent} & V2V &  &  & \checkmark & \checkmark &  &  &7,788  & 2-10&\href{https://github.com/YuanYunshuang/FPV_RCNN}{Link}\\
            OPV2V~\cite{xu2022opv2v} & ICRA 2022 & CARLA \& OpenCDA~\cite{xu2021opencda} & V2V & \checkmark &  & \checkmark & \checkmark  & \checkmark  & \checkmark  & 11,464 & 2-7&\href{https://mobility-lab.seas.ucla.edu/opv2v}{Link}\\
            AUTOCASTSIM~\cite{cui2022coopernaut} & CVPR 2022 & CARLA~\cite{dosovitskiy2017carla} & V2V & \checkmark &  & \checkmark & \checkmark &  &  &  -& -&\href{https://ut-austin-rpl.github.io/Coopernaut}{Link}\\
            DAIR-V2X-C~\cite{yu2022dair} & CVPR 2022 & Real-World & V2I & \checkmark &  & \checkmark & \checkmark &  &  & 38,845 & 2&\href{https://thudair.baai.ac.cn/cooptest}{Link}\\
            V2XSet~\cite{xu2022v2x} & ECCV 2022 & CARLA\cite{dosovitskiy2017carla} \& OpenCDA~\cite{xu2021opencda} & V2V,V2I & \checkmark &  & \checkmark &\checkmark  &  &  & 11,447 & 2-5&\href{https://github.com/DerrickXuNu/v2x-vit}{Link}\\
            DOLPHINS~\cite{mao2022dolphins} & ACCV 2022 & CARLA~\cite{dosovitskiy2017carla} & V2V,V2I & \checkmark &  & \checkmark & \checkmark & \checkmark &  & 42,376 & 3&\href{https://dolphins-dataset.net/}{Link}\\
            CARTI~\cite{bai2022pillargrid} & ITSC 2022 & CARLA~\cite{dosovitskiy2017carla} & V2I &  &  & \checkmark & \checkmark &  &  & 11,000 & 2&-\\
            V2V4Real~\cite{xu2023v2v4real} & CVPR 2023 & Real-World & V2V &\checkmark &  & \checkmark& \checkmark & \checkmark & &20,000 & 2&\href{https://research.seas.ucla.edu/mobility-lab/v2v4real}{Link}\\
            
            V2X-Seq (SPD)~\cite{yu2023v2x} & CVPR 2023 & Real-World & V2V,V2I & \checkmark & & \checkmark & & \checkmark & & 15,000& 2 & \href{https://github.com/AIR-THU/DAIR-V2X-Seq}{Link}\\

            DeepAccident~\cite{wang2023deepaccident} & - & CARLA\cite{dosovitskiy2017carla} & V2V,V2I & \checkmark & & \checkmark & \checkmark & \checkmark & \checkmark & 57,000& 5 & \href{https://deepaccident.github.io/}{Link}\\
		\hline
	\end{tabular}
    }
      }
   \end{center}
    \vspace{-4mm}
   \label{tab:diff-datasets}
\end{table*}
Since the development of autonomous vehicles is still in the early stages, there are a large number of L0-L2 CVs in the road network, which lack advanced perception and planning capabilities and only provide driving assistance. These CVs still possess the ability to connect to the V2X network through OBU, telematics-box (T-BoX), and APP. Note that T-BoX is another type of communication device with functions similar to OBU.
Besides, the controller area network (CAN) is a reliable and cost-effective communication protocol, which enables data exchange and sharing among multiple nodes, supporting real-time control and response. 
CAN acts as a bridge, transmitting the received messages to the human-machine interface (HMI) and advanced driver assistance systems (ADAS).
HMI in CVs consists of infotainment systems, vehicle diagnostics, \etc, which can provide drivers with real-time information.
ADAS can alert about potential hazards such as accidents, roadworks, and traffic congestion, and also enable intelligent navigation, considering real-time traffic conditions.
HMI and ADAS can improve safety and reduce the risk of accidents.

\section{Datasets}
\label{sec:datasets}
Large datasets are essential for training generalized and robust perception models. However, most perception datasets are mainly built for individual perception scenarios, not practical for cooperative perception. In this way, a significant amount of effort is put into building collaborative perception (CP) datasets that are customized for V2X scenarios. We summarize the existing cooperative perception datasets with detailed information as shown in Table~\ref{tab:diff-datasets}.

Acquiring real-world V2X perception data and building benchmarks is considerably more challenging than other perception tasks that can readily access data from the internet, mainly due to the exorbitant costs.
From Table~\ref{tab:diff-datasets}, we observe that most CP datasets are built based on simulation, making the expenses acceptable. Also, constructing datasets from the real world has also gotten increasing attention in recent years, as it eliminates the challenge of bridging the gap between simulation and the real world. 
The most popular scenario for cooperative perception is V2V (vehicle-to-vehicle), while few datasets~\cite{li2022v2x,xu2022v2x,mao2022dolphins} are available that support both V2V and V2I (vehicle-to-infrastructure). Like individual perception, detection is also the most popular task for cooperative perception. 

Furthermore, the viewpoint represents the number of collaborating agents in a single frame, and most datasets support fewer than $5$ collaborating agents, resulting in a sparse distribution of collaborating agents over a large spatial range. However, in future V2X scenarios, there are likely to be a large number of connected vehicles. Therefore, a promising direction is to build datasets with a larger number of collaborating agents in a single frame, achieving a dense distribution of collaborating agents on the road and better alignment with future scenarios.
In the following paragraph, we introduce simulators, CP datasets based on simulation and real-world respectively, and end with an introduction of the CP tasks.

\subsection{Simulators}
A traffic simulator is a software tool that allows researchers to evaluate algorithms in virtual environments, where real-world traffic scenarios (\eg, road layouts and traffic flow) are simulated. Researchers can leverage the simulator to generate large amounts of data for training and testing under various scenarios and conditions without the need for expensive physical testing on real roads.

\textbf{CARLA}~\cite{dosovitskiy2017carla} is an advanced and highly versatile open-source simulator designed for autonomous driving research and development. One of its notable strengths lies in the integration with the powerful Unreal Engine, which provides exceptional visual fidelity and realistic scene rendering.
CARLA provides a range of built-in sensors, including cameras, LiDARs, and radars, enabling users to generate sensor data that closely emulates real-world perception systems. 
However, the behavior rules employed by CARLA's traffic manager are simplistic and do not fully capture the complexities of real driver behavior.

\textbf{SUMO}~\cite{krajzewicz2012recent} (Simulation of Urban Mobility) is a widely-used open-source traffic and driver behavior simulator. It offers a powerful platform for simulating large-scale, realistic traffic scenarios. SUMO's primary strength lies in its ability to handle complex traffic flows and accurately represent driver behavior.
SUMO takes into account factors such as vehicle acceleration, deceleration, lane changing, and interaction with other vehicles, enabling a more accurate portrayal of real-world traffic scenarios.

In summary, SUMO is a powerful open-source traffic and driver behavior simulator that complements CARLA's capabilities in the autonomous driving framework. 

\subsection{Datasets based on Simulation}
Due to the difficulty in retrieving data from the real world, many datasets of collaborative perception are generated in simulators. Conventional simulators include CARLA~\cite{dosovitskiy2017carla}, SUMO~\cite{krajzewicz2012recent}, OpenCDA~\cite{xu2021opencda}, \etc~
Collecting data from the simulator is economical in time and budget. However, a domain gap exists between the simulator and the real world, thus some work put efforts to narrow such a gap. Here we introduce three popular datasets based on simulation.

\textbf{V2X-Sim}~\cite{li2022v2x} employs SUMO to simulate traffic flows and CARLA to collect sensor streams. Retrieving data from the roadside unit and vehicles, V2X-Sim supports both V2V and V2I scenarios. Besides, V2X-Sim provides a benchmark for three tasks, \ie, object detection, multi-object tracking, and semantic segmentation. The whole dataset is divided as $8,000$/$1,000$/$1,000$ frames for training/validation/testing, with $2$-$5$ collaborative vehicles in each scene. 


\textbf{V2XSet}~\cite{xu2022v2x} is a large-scale V2X perception dataset based on CARLA and OpenCDA. Compared to previous datasets, V2XSet supports the simulation of localization error and time delay, which is more approaching to real-world settings. V2XSet has 11,447 frames in total, where train/validation/test splits are $6,694$/$1,920$/$2,833$ frames respectively. Besides, V2XSet covers $5$ types of roadway, \ie, straight segment, curvy segment, midblock, entrance ramp, and intersection. In each scene, there are $2$-$7$ intelligent agents for collaborative perception.

\subsection{Datasets based on Real-world Scenarios}
With the pursuit of more realistic data, benchmarks that are built upon the real world get increasing attention. Compared to simulation, real-world scenarios showcase more complicated traffic behaviors and noisy data, which is beneficial for obtaining robust CP models. Here we introduce two datasets based on real-world scenarios.

\textbf{DAIR-V2X}~\cite{yu2022dair} is the first large-scale, real-world dataset for V2I cooperative perception. DAIR-V2X is composed of three parts, \ie, DAIR-V2X-C, DAIR-V2X-I, and DAIR-V2X-V, where DAIR-V2X-C contains sensor information from both vehicle and infrastructure. DAIR-V2X-C has $38,845$ camera frames and $38,845$ LiDAR frames. Focusing on the V2I 3D object detection problem (VIC3D), DAIR-V2X-C records nearly $464,000$ 3D bounding boxes for $10$ classes.

\textbf{V2V4Real}~\cite{xu2023v2v4real} is the first large-scale real-world multi-modal dataset for V2V perception. V2V4Real has $20$K LiDAR frames and $40$K RGB frames, with $240$K annotated 3D bounding boxes for $5$ classes. Four road types are included in V2V4Real, \ie, intersection, highway entrance ramp, highway straight road, and city straight road, which are captured in Columbus, Ohio, in the USA. V2V4Real supports three cooperative perception tasks, \ie, 3D object detection, 3D object tracking, and  Sim2Real domain adaptation.

\subsection{Task and Evaluation}
In this section, we introduce popular collaborative perception tasks in V2X scenarios, \ie, collaborative 3D object detection and semantic segmentation, including their definitions and metrics for evaluation.

\textbf{3D object detection} aims to produce bounding boxes for objects and recognize their categories in 3D scenes. Basically, a 3D bounding box can be denoted as $(x,y,z,l,w,h,\theta)$, where $(x,y,z)$ represents the center, $(l,w,h)$ represents the size, $\theta$ is the heading angle of the bounding box~\cite{mao20223d}. 3D object detection can also be conducted under the Bird's Eye View (BEV), where 2D bounding boxes are utilized to locate objects. The most common metric for 3D object detection is Average Precision (AP), which is defined as the area below the precision-recall curve. When detected objects are in more than one category, mean Average Precision (mAP) is calculated as the average AP of all categories.

\textbf{Semantic segmentation} aims to predict pixel-level categories for RGB images or point-level categories for point clouds. Collaborative semantic segmentation is generally conducted under the BEV. Mean Intersection over Union (mIoU) between the prediction and the ground truth is utilized as the performance metric, which is averaged across categories. Besides, mean Accuracy (mAcc) is also widely used, where accuracy is the ratio of correctly categorized pixels/points over all pixels/points.

\section{Taxonomy}
\label{sec:taxonomy}
In this section, we summarize and categorize the existing collaborative perception methods from various aspects, where numerous works put significant efforts to address critical issues to drive the rapid development of this field.
\subsection{Collaboration Stage}
\begin{figure}
   \begin{center}
      \includegraphics[width=0.9\linewidth]{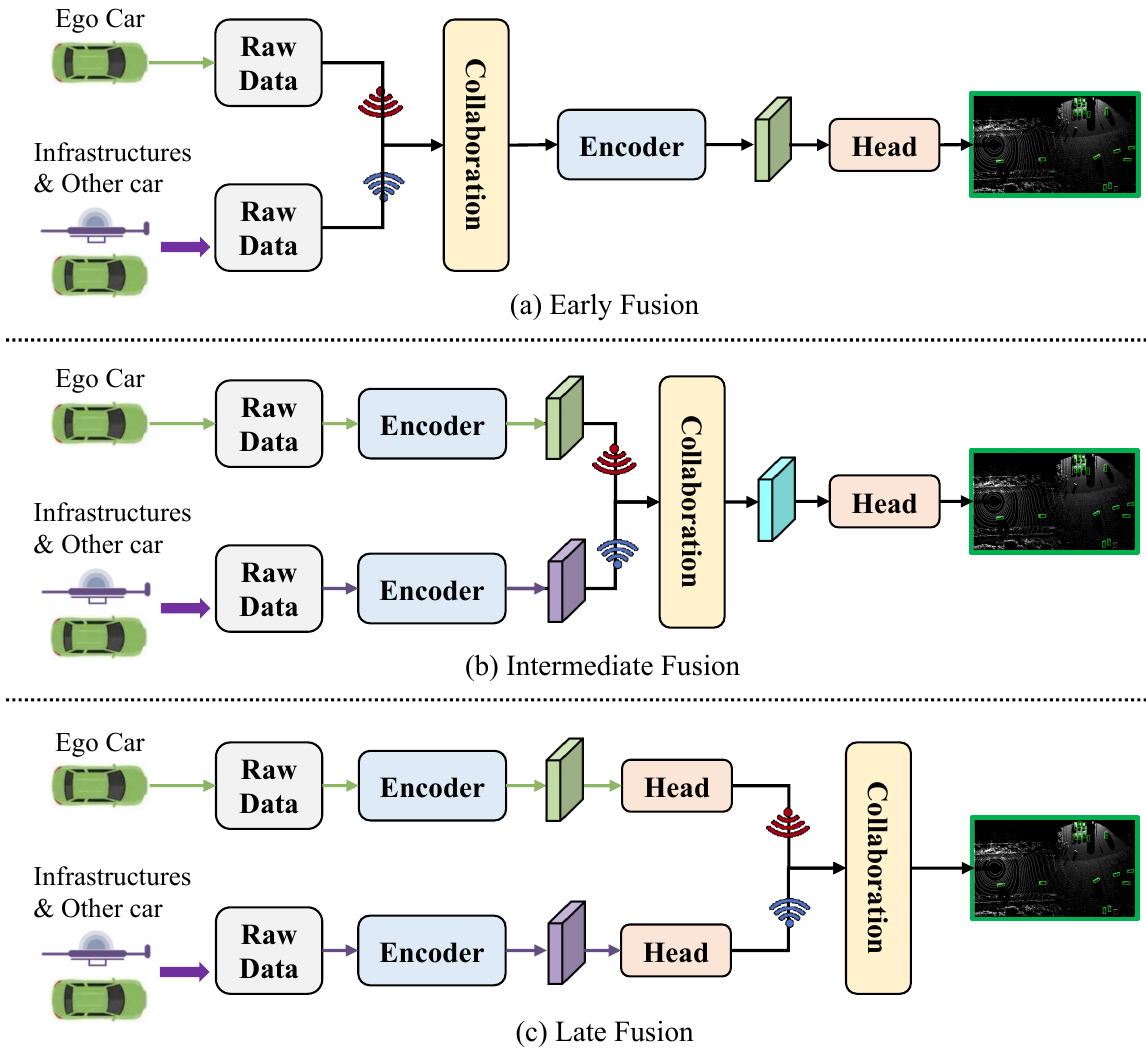}
   \end{center}
   \vspace{-5mm}
      \caption{Illustration of different collaboration stage.}
      \vspace{-4mm}
   \label{fig:fusion-stage}
\end{figure}

The objective of V2X perception is to detect objects in traffic environments using sensors on vehicles and other devices, which presents a multi-agent sensor fusion problem.
V2X perception can be broadly classified into three types according to the fusion stage, \ie, early, intermediate, and late fusion, as shown in Fig.~\ref{fig:fusion-stage}. 

\textbf{Early Fusion}~\cite{gao2018object,chen2019cooper,arnold2020cooperative} directly transforms raw data and merges it to form a comprehensive perception in the processing pipeline.
In the system that uses both cameras and LiDAR sensors, the raw data from each sensor can be fused at the pixel level and point cloud level, respectively.
This means that the raw images from the cameras and the point clouds from the LiDAR sensors are combined to form a single representation of the environment.
To achieve this, early fusion techniques often involve pre-processing and calibration steps to align and normalize the data from different sensors.
For example, ~\cite{gao2018object} unsample and convert the point clouds into pixel-level depth information, which is then connected with RGB images.
~\cite{chen2019cooper} applies sparse convolution to support detection in low-density point cloud data. 
~\cite{arnold2020cooperative} proposes to combine point clouds from different sensing points to improve 3D object detection, which involves transmitting each point cloud to a central fusion system, where they are concatenated into a single point cloud and input into the detection model.
The benefits of early fusion include the ability to integrate information from multiple sensors at a low level, reducing the complexity of subsequent processing stages and providing more information for downstream models.
However, early fusion can also be challenging due to the differences in the data generated by different sensors, such as different resolutions, laser beams, and noise levels.

\textbf{Intermediate Fusion}~\cite{chen2017multi,wang2020v2vnet,li2021learning,pmlr-v155-vadivelu21a,liu2020when2com,lei2022latency,xu2022opv2v,xu2022v2x,Su2022uncertainty,huwhere2comm,xu2022cobevt,cui2022coopernaut,liu2020who2com,yu2023vehicle,xu2022bridging,marvasti2020cooperative,lu2023robust,bai2022pillargrid,guo2021coff,qiao2023adaptive,yu2023vehicle,meng2023hydro,zhang2022multi,bai2023cooperverse,wang2023core,yang2023spatio,su2023collaborative,li2023multi} is the primary scheme in CP model design, where ego vehicle's perception model integrates its own intermediate features with intermediate features of other collaborative agent's models, achieving collaborative perception. Intermediate collaboration has drawn the attention of researchers due to its balance between accuracy and transmission bandwidth. A deep fusion scheme is proposed by~\cite{chen2017multi} to combine region-wise features from multiple views, which enables interactions between intermediate layers of different paths.~\cite{wang2020v2vnet} proposes a spatial-aware message-passing mechanism to jointly reason detection and prediction. Learning-based methods~\cite{pmlr-v155-vadivelu21a,li2021learning} aim to reduce vehicle localization errors and improve training.~\cite{liu2020when2com} proposes a communication framework to avoid unnecessary transmissions between connected cars, reducing communication bandwidth.~\cite{lei2022latency} takes information transmission latency into consideration.~\cite{xu2022opv2v} guides the network to pay attention to key observations according to interactions of features from neighboring connected vehicles.~\cite{xu2022v2x} presents a unified transformer architecture for V2X perception that can capture the heterogeneous nature of V2X systems and provide robustness against various noises.~\cite{huwhere2comm} calculates a confidence score for each spatial area of the feature, electing perceptually critical regions for sharing.~\cite{xu2022cobevt} generates BEV segmented maps with multi-agent multi-camera sensors. 

\textbf{Late Fusion}~\cite{zeng2020dsdnet,melotti2020multimodal,fu2020depth, caltagirone2019lidar,2012Car2X,shi2022vips,glaser2023we,chen2022learning} is a practical technique used in V2X collaborative perception, where the perception results from different agents are combined.
In late fusion, the individual sensor outputs are processed separately, and the results are combined at the decision level to create a final perception.
Although early fusion provides essential contextual information that late fusion cannot provide, it requires significant transmission bandwidth. 
The advantages of late fusion include the ability to incorporate information from different models without extra training, and the flexibility to integrate the results from different data modalities.
However, late fusion can also be challenging due to the need for accurate alignment of the individual outputs and the potential for increased complexity in the result-fusing process. Also, late fusion suffers from the drawback of error accumulation.

\subsection{Roadside Visual Sensors}
Roadside visual sensors are critical components for V2X collaborative perception, including RGB cameras, Radar, and LiDARs. For example, LiDAR provides reliable geometry information and can be directly transferred to real-world coordinate systems.  
In addition, roadside sensors are typically installed at a high position along the roadside, providing vehicles with an extra wide field of view.

Unlike onboard visual sensors on the vehicle, the placement of roadside sensors has greater flexibility.
Therefore, in V2X scenarios, an important problem is how to determine the optimal placement for roadside sensors to maximize their benefits since different positions bring different views.
Cai \etal~\cite{cai2022analyzing} propose a LiDAR simulation library for the simulation of infrastructure LiDAR sensors. They also analyze the correlation between perception performance and the density/uniformity of the LiDAR point cloud.
Jiang \etal~\cite{cai2022analyzing} propose a perceptual gain-based greedy algorithm that obtains approximate optimal solutions for roadside LiDAR placement optimization and introduce a perception predictor that can quickly obtain the perceptual gain by predicting the perception ability of a LiDAR placement.

There are also works focusing on improving the performance of infrastructure-only perception. ~\cite{yang2023bevheight} refines the depth estimation method of BEV roadside sensors to support camera-based detection. ~\cite{zimmer2023infradet3d} utilizes early fusion to fuse point clouds from two roadside LiDARs. Besides, a monocular camera is incorporated to improve robustness. ~\cite{shi2023center} proposes a center-based detector for Roadside LiDARs. 

\subsection{Communication Latency Compensation}
Communication latency (time delay) among CAVs presents an inevitable challenge in the practical application of V2X, as it impairs perception performance and even leads to worse performance than individual perception. To address this issue, numerous works design methods to compensate for the performance degradation caused by communication latency and exhibit greater robustness to the latency. 
Overall, these methods can be classified into two categories: (\romannumeral1) Some methods (\eg, SyncNet~\cite{lei2022latency}) adopt historical information and transform the time delay compensation problem into a temporal prediction problem. (\romannumeral2) Some methods design implicit delay-aware modules trained on data with simulated communication latency or develop specific transmission strategies to reduce the latency impact on performance, such as AutoCast~\cite{qiu2021autocast}, AVR~\cite{qiu2018avr}, V2X-ViT~\cite{xu2022v2x}, and DAIR-V2X~\cite{yu2022dair}.

Early fusion methods design transmission strategies to alleviate the latency. AutoCast~\cite{qiu2021autocast} decides which objects to transmit at every decision interval and in what order, which is treated as a Markov decision process. 
The algorithm selects the set of transmitted objects at each time step based on the current environment. 
AVR~\cite{qiu2018avr} estimates, at the sender-side, the motion vector of dynamic objects by using homography matrix and optical flow segmentation, and uses the motion vector to estimate the current position and compensate for the delay at the receiver-side.

Intermediate fusion methods adopt data-driven implicit compensation modules. In V2X-ViT~\cite{xu2022v2x}, an adaptive delay-aware positional encoding module (DPE) is designed for time alignment, compensating for latency by inputting delay time information. Instead, SyncNet~\cite{lei2022latency} designs a temporal prediction module, which adopts the feature-attention symbiotic estimation (FASE) methods to estimate the missing collaboration features and introduces time modulation (TM) to balance the noise. It can work as a plugin latency compensation module for other intermediate collaborative methods. 
One advantage of these methods is they can be optimized end-to-end. Note that the complex models may also bring additional noise and reduce performance.

For late fusion methods, DAIR-V2X~\cite{yu2022dair} proposes the Time Compensation Late Fusion (TCLF) framework, which involves the boxes matching between successive frames, estimation of velocity, linear interpolation to approximate vehicle positions, and ultimately fusing the results.

\subsection{Collaboration Efficiency}
The collaborative perception system improves perception performance by transferring complementary information between different agents (vehicles or infrastructure). However, the communication bandwidth is usually limited in real application scenarios. Therefore, the collaborative perception system faces an inevitable and fundamental challenge, \ie, how to balance the trade-off between perception performance and communication bandwidth. 
Most works~\cite{xu2022v2x,li2021learning,xu2022cobevt} reduce the communication cost by simply leveraging $1\times 1$ convolutions to compress and decompress the transmission features along the channel dimension. In this way, although the bandwidth is reduced, the performance also decreases accordingly. 

Several works leverage learnable methods to facilitate collaboration efficiency.
V2VNet~\cite{wang2020v2vnet} first adopts a LiDAR backbone to process the input point cloud and obtains a BEV feature. Then, it utilizes the variational image compression algorithm~\cite{ballevariational} to compress the feature achieving the transmission feature.
DiscoNet~\cite{li2021learning} designs a teacher-student distillation method.
Through feature distillation and proposed matrix-valued edge weight for highlighting the informative regions, DiscoNet can promote a better performance-bandwidth trade-off.
Who2com~\cite{liu2020who2com} proposes a three-stage communication mechanism: request, match, and connect in order to select the best matching agents for communication. 
When2com~\cite{liu2020when2com} utilizes scaled general attention for the ego agent to decide when to communicate. It treats the collaborative perception problem as learning to construct the communication group and to decide when to communicate without explicit supervision for such decisions during training.
FPV-RCNN~\cite{fpvrcnn} adopts a two-stage framework. It first generates object proposals and then selects key-points in those proposals for feature communication. The feature selection process reduces the redundancy of collaborative features and thus decreases the bandwidth requirement.
Where2comm~\cite{huwhere2comm} proposes a spatial-confidence-aware communication strategy by learning a spatial confidence map to reflect the perceptually critical areas in the transmission features. According to such a confidence map, each agent can transfer spatially sparse yet critical features to partners, achieving higher performance with less communication bandwidth.
Selective Communication~\cite{chiu2023selective} designs a 2-phase communication procedure. In the first phase, connected agents share the lightweight self-detection results (\ie, centers of the detected objects) with each other to determine whom to communicate with. Then point features are transmitted in the second phase. 
Lin \etal~\cite{lin2023your} propose an image-aided point clouds compression framework, where depth information is estimated from images and utilized to guide the compression for point clouds. 
UMC~\cite{wang2023umc} introduces a two-stage entropy-based selection module to select appropriate regions for communication.

\subsection{Adversarial Attack and Defense}
As is widely recognized, the robustness of deep neural networks presents significant challenges, particularly in the context of adversarial attacks, and its security is difficult to ensure. Moreover, collaborative perception systems, with their communication and fusion modules, face inherent challenges in defending against adversarial attacks because of malicious or unreliable shared information. Therefore, ongoing efforts are being made to address these challenges and enhance the security of collaborative perception systems.

Tu \etal~\cite{tu2021adversarial} investigates adversarial attacks on multi-agent communication. Experiments show that perturbed transmissions can cause performance degradation. A cooperative system with a larger number of benign agents can effectively mitigate the impact of adversarial attacks.
To mitigate these attacks, they directly employ adversarial training. Although it is effective, it comes with the drawback of introducing extra overhead during training and an inability to generalize to unseen attackers.

Li \etal~\cite{li2023among} develop a defense strategy named ROBOSAC, which demonstrates exceptional ability to generalize unseen attackers. 
Instead of adversarial training with all available information, ROBOSAC leverages an intelligent selection of collaborators. 
The underlying assumption is that collaborative messages from attackers can cause a large number of divergences while messages from friendly collaborators often reach a consensus with ego-car. 
Given the attacker ratio and a sampling budget, ROBOSAC derives the maximum expected number of benign collaborators, achieving a trade-off between performance and complexity. 

V2XP-ASG~\cite{xiang2022v2xp} is an adversarial scene generator, which can produce challenging scenes for LiDAR-based methods. In detail, V2XP-ASG can construct an adversarial collaboration graph and perturb agents' poses in an adversarial and plausible way. Experiments show that training on the generated challenging scenes can further improve the performance of collaborative perception systems.

\subsection{Pose Alignment}
CP methods require sensor poses of each agent to transform raw data, intermediate features, or final predictions to ego-car's coordinates. However, pose/localization errors are commonly caused by satellite positioning noises and asynchronous between collaborative agents. Inaccurate sensor poses may cause misalignment during the information fusion stage and cause significant performance degradation, even below single-vehicle performance under realistic noise. To alleviate this problem, several works propose specially designed pose alignment modules.

For intermediate fusion methods, Vadivelu \etal~\cite{robustv2vnet} propose end-to-end learnable neural reasoning layers to estimate pose errors and reach a consensus about the errors. 
Specifically, the pose regression module outputs a pose correction perturbation and produces a predicted true relative transformation accordingly. Then, the consistency module refines the relative pose estimates from the regression module by finding a set of globally consistent absolute poses among all agents with Markov random field (MRF). Finally, to focus on clean messages and ignore noisy ones, they propose a simple yet effective attention mechanism to assign a weight to each warped message before they are averaged. 
Yuan \etal~\cite{fpvrcnn} propose a localization error correction module to alleviate the performance deterioration. The module first selects keypoints of poles, fences and walls according to the classification scores and then uses the maximum consensus algorithm~\cite{chin2017maximum} with a rough searching resolution to find the corresponding vehicles centers and poles points. It finally uses the correspondence to estimate accurate pose error.
Lu \etal~\cite{lu2023robust} propose CoAlign, a novel hybrid collaboration framework that is robust to unknown pose errors. It utilizes a novel agent-object pose graph modeling similar to graph-based SLAM to enhance pose consistency among collaborating agents. It doesn’t require any ground-truth pose supervision and is thus more practical.

For late fusion methods, Song \etal~\cite{song2023cooperative} propose OptiMatch, where a transport theory-based algorithm is devised to construct an associated co-visible object set, and a correction transform is estimated from the matched object pairs and further applied to the noisy relative transform. It finally performs global fusion and dynamic mapping.

\begin{table*}
   \caption{Summary and comparison of essential characteristics for different CP methods on OPV2V dataset.
   }
   \vspace{-7mm}
   \begin{center}
      \resizebox{\linewidth}{!}{
         \renewcommand\arraystretch{1.3}
    \footnotesize
    \setlength{\tabcolsep}{3.5pt}{
        \begin{tabular}{c|cccccccccc}
		\hline
            Method & Publication & Collab Scheme & Backbone & Param (M) & MACs (G) & Transmission (MB) & Inference Time (ms) \\ 
		\hline
            Late Fusion & - & Late & PointPillars & 6.58 & 31.34 & - & 5.83 \\
            Cooper~\cite{chen2019cooper} & ICDCS 2019 & Early & PointPillars & 6.58 & 31.83 & - & 7.08 \\
		  F-Cooper~\cite{fcooper} & SEC 2019 & Intermediate & PointPillars & 7.27 & 144.36 & 6.29 & 24.78 \\
            V2VNet~\cite{wang2020v2vnet} & ECCV 2020 & Intermediate & PointPillars & 14.61 & 447.28 & 12.58 \\
            OPV2V~\cite{xu2022opv2v} & ICRA 2022 & Intermediate & PointPillars & 6.58 & 64.20 & 11.01 & 15.78 \\
            CoBEVT~\cite{xu2022cobevt} & CoRL 2022 & Intermediate & PointPillars & 10.49 & 194.34 & 6.29 & 56.50 \\
            FPV-RCNN~\cite{fpvrcnn} & RAL 2022 & Intermediate (2-stage) & PV-RCNN & 2.89 & 128.12 & - & - \\
            
		\hline
	\end{tabular}
    }

      }
   \end{center}
   \label{tab:methods-opv2v}
   \vspace{-3mm}
\end{table*}

\begin{table*}
   \caption{Summary and comparison of essential characteristics for different CP methods on V2XSet dataset.}
   \vspace{-7mm}
   \begin{center}
      \resizebox{\linewidth}{!}{
         \renewcommand\arraystretch{1.3}
    \footnotesize
    \setlength{\tabcolsep}{3.5pt}{
        \begin{tabular}{c|ccccccccccc}
		\hline
            Method & Publication & Collab Scheme & Backbone & Param (M) & MACs (G) & Transmission (MB) & Inference Time (ms) \\
		\hline
            Late Fusion & - & Late & PointPillars & 6.58 & 31.34 & - & 5.52 \\
            Early Fusion & - & Early & PointPillars & 6.58 & 31.45 & - & 6.29 \\
            F-Cooper~\cite{fcooper} & SEC 2019 & Intermediate & PointPillars & 7.27 & 144.36 & 6.29 & 17.35 \\
            V2VNet~\cite{wang2020v2vnet} & ECCV 2020 & Intermediate & PointPillars & 14.61 & 447.28 & 12.58 & 66.54 \\
            DiscoNet~\cite{mehr2019disconet} & NeurIPS 2021 & Intermediate & PointPillars & - & - & - & - \\
            OPV2V~\cite{xu2022opv2v} & ICRA 2022 & Intermediate & PointPillars & 6.58 & 64.20 & 11.01 & 11.38 \\
            CoBEVT~\cite{xu2022cobevt} & CoRL 2022 & Intermediate & PointPillars & 10.49 & 194.34 & 6.29 & 53.04 \\
            V2X-ViT~\cite{xu2022v2x} & ECCV 2022 & Intermediate & PointPillars & 12.46 & 229.74 & 6.29 & 94.75 \\
            Where2comm\cite{huwhere2comm} & NeurIPS 2022 & Intermediate & PointPillars & 8.06 & 193.63 & 11.01 & 25.95 \\
		\hline
	\end{tabular}
    }

      }
   \end{center}
   \label{tab:methods-v2xset}
   \vspace{-3mm}
\end{table*}

\begin{table*}
   \caption{Summary and comparison of essential characteristics for different CP methods on V2X-Sim dataset. 
   }
   \vspace{-7mm}
   \begin{center}
      \resizebox{\linewidth}{!}{
         \renewcommand\arraystretch{1.3}
    \footnotesize
    \setlength{\tabcolsep}{3.5pt}{
        \begin{tabular}{c|cccccccccc}
		\hline
            Method & Publication & Collab Scheme & Backbone & Param (M) & MACs (G) & Transmission (MB) & Inference Time (ms) \\
		\hline
            Late Fusion & - & Late & FaF & 7.90 & 94.00 & - & 19.82 \\ 
            Early Fusion & - & Early & FaF & 7.90 & 94.00 & - & 19.87 \\
            Who2com~\cite{liu2020who2com} & ICRA 2020 & Intermediate & FaF & 20.28 & 186.01 & 1.05 & 39.88 \\
            When2com~\cite{liu2020when2com} & CVPR 2020 & Intermediate & FaF & 20.28 & 186.01 & 1.05 & 39.56 \\
            V2VNet~\cite{wang2020v2vnet} & ECCV 2020 & Intermediate & FaF & 7.90 & 94.00 & 3.15 & 58.31 \\
            DiscoNet~\cite{mehr2019disconet} & NeurIPS 2021 & Intermediate & FaF & 7.97 & 95.81 & 1.05 & 47.59 \\
            
		\hline
	\end{tabular}
    }
    
      }
   \end{center}
   \label{tab:methods-v2xsim}
   \vspace{-3mm}
\end{table*}
\subsection{Others}
\textbf{Heterogeneous Model Fusion.} Most CP methods use the same perception models. Such a setting is understandable since different models may injure the performance due to mismatches in architectures and parameters. However, in real scenarios, cars are likely to be equipped with different co-perception models since different companies may adopt different models.~\cite{chen2022model} proposes a model-agnostic multi-agent perception framework. 
To align confidence distributions of different models, a calibrator is trained for each model independently. 
A Promote-Suppress Aggregation (PSA) is introduced to handle bounding box aggregation. 
Thus~\cite{chen2022model} improves the performance of collaborative with agnostic models, yet are only suitable for late fusion. ~\cite{xiang2023hm} proposes the first unified multi-agent hetero-modal framework HM-ViT, supporting collaborative perception with varying numbers and types of vehicles.~\cite{bai2022vinet} designs Two-Stream Fusion Module to process the  heterogeneous fusion between vehicles and infrastructures.

\textbf{Camera-only Cooperative Perception.} Though most works focus on cooperative perception under LiDAR-only or LiDAR-camera fusion settings, there are works taking insights into camera-only cooperative perception. Wang \etal~\cite{wang2023vimi} introduce an intermediate fusion method for camera-based V2I scenarios. Considering the same object is detected by vehicle and infrastructure at different distances, the features extracted by vehicle and infrastructure can be different in scale. To fuse multi-scale features, a Multi-scale Cross Attention (MCA) is designed, where features of different scales are obtained by MeanPooling operation first, then cross-attention is utilized to pick out matched features. Also, Camera-aware Channel Masking (CCM) technique takes camera parameters as priors to augment image features.
Hu \etal~\cite{hu2023collaboration} introduce another work that investigates camera-only 3D detection. Basically,~\cite{hu2023collaboration} proposes Collaborative Depth Estimation (Co-Depth) to improve the accuracy of depth estimation. In Co-Depth, agents share key depth information to refine the estimated depth probabilities. A Collaborative Detection Feature Learning (Co-FL) technique allows agents to select interesting parts of the BEV feature to share with each other, which saves communication costs. 

\textbf{Unsupervised Learning.}
~\cite{han2023ssc3od} proposes a sparsely supervised collaborative 3D object detection framework SSC3OD. Specifically, SSC3OD first utilizes Pillar-MAE module to reason over high-level semantics in an unsupervised way, then pseudo labels are generated by the instance mining module. ~\cite{chen2022co} utilizes point clouds from vehicle-side and infrastructure-side to conduct contrastive learning. Besides, ~\cite{chen2022co} proposes contextual shape prediction as a pre-training task for unsupervised representation learning.

\textbf{Federated Learning.}
~\cite{tian2022federated} proposes federated vehicular Transformers to apply privacy-preserving computing in cooperation systems. The core of federated vehicular Transformers is secure cross-vehicle attention layers, which adopt privacy-preserving techniques (\eg, homomorphic encryption, secret sharing) to encrypt multi-modal features and fuse them in a privacy-preserving way. To overconme degradation caused by dynamic states of connected vehicles in federated learning, ~\cite{song2023v2x} designs a contextual client selection pipeline. In detail, ~\cite{song2023v2x} pre-clusters clients based on the similarity of local data distribution, and selects clients with minimal latency for future fusion.

\textbf{Lossy Communication.}
Li \etal~\cite{li2023learning} propose a framework to tackle lossy communication problems. The LC-aware Repair Network is introduced to repair the damaged areas of the input features. Besides, V2V Attention Module is utilized to fuse the features from various sources, considering both intra-vehicle attention and inter-vehicle attention. 
Besides, Ren \etal~\cite{ren2023interruption} utilize historical information to repair missing information caused by lossy communication.~\cite{ren2023interruption} proposes an adaptive prediction model to generate and select promising features for the lossy areas.

\textbf{Sim-to-Real Generalization.}
Li \etal~\cite{li2023s2r} proposes S2R-ViT to transfer simulation to reality for cooperative perception. Specifically, an uncertainty-aware transformer is utilized to overcome time latency and localization errors in real-world data. Besides, discriminators are designed to classify whether a feature is real or simulated, helping the model generate domain-invariant representations.

\section{Experimental Analysis}
\label{sec:exps}
In this section, we present extensive quantitative and qualitative experimental analyses to examine current collaborative perception (CP) methods in V2X scenarios. 
In \S\ref{subsec:compare}, we provide detailed model attribute comparisons on various datasets. In \S\ref{subsec:trade-off}, we analyze and discuss the performance-bandwidth trade-off issue. 
In \S\ref{subsec:range}, we investigate the effect of different LiDAR ranges.
From \S\ref{subsec:time} to \S\ref{subsec:mix}, we investigate the model robustness against various noises, including communication latency, localization errors, lossy communication, and mixed noises, where noises are generated to simulate the real-world application environment. 
In \S\ref{subsec:sim2real}, we examine the sim-to-real generalization ability of existing CP methods. 
In \S\ref{subsec:qualitative}, we show the qualitative comparisons.

We perform experiments on several datasets including simulation-based datasets, \ie, OPV2V~\cite{xu2022opv2v}, V2X-Sim~\cite{li2022v2x}, and V2XSet~\cite{xu2022v2x}, and real-world based datasets, \ie, DAIR-V2X~\cite{yu2022dair} and V2V4Real~\cite{xu2023v2v4real}. Concretely, we adopt OpenCOOD\footnote{\href{https://github.com/DerrickXuNu/OpenCOOD}{https://github.com/DerrickXuNu/OpenCOOD}} as the codebase for OPV2V and V2XSet datasets. We adopt CoPerception\footnote{\href{https://github.com/coperception/coperception}{https://github.com/coperception/coperception}} and V2V4Real\footnote{\href{https://github.com/ucla-mobility/V2V4Real}{https://github.com/ucla-mobility/V2V4Real}} as the codebases for the V2X-Sim~\cite{li2022v2x} and V2V4Real~\cite{xu2023v2v4real} datasets, respectively. Note that, on the DAIR-V2X dataset, results of some methods are produced based on OpenCOOD, and for those methods that are not supported by OpenCOOD, we use Where2comm\footnote{\href{https://github.com/MediaBrain-SJTU/Where2comm}{https://github.com/MediaBrain-SJTU/Where2comm}} as the codebase to conduct experiments.

All the models involved in the experimental analysis were first trained from scratch in a \textit{perfect} setting, \ie, without any noise and compression, using the default hyperparameters of corresponding codebases, with the exception of disabling the multi-scale feature. Then the pre-trained models are fine-tuned under different settings for specific analyses.
We select checkpoints for evaluation according to the validation loss.
Note that the V2V4Real dataset has data augmentation enabled by default.

\begin{figure*}[htbp]
    \centering
    \captionsetup[subfloat]{font=scriptsize}
    \subfloat[DAIR-V2X-C]{
        \label{fig:compress_dair}
        \includegraphics[height=0.295\textheight,trim=12 10 20 10,clip]{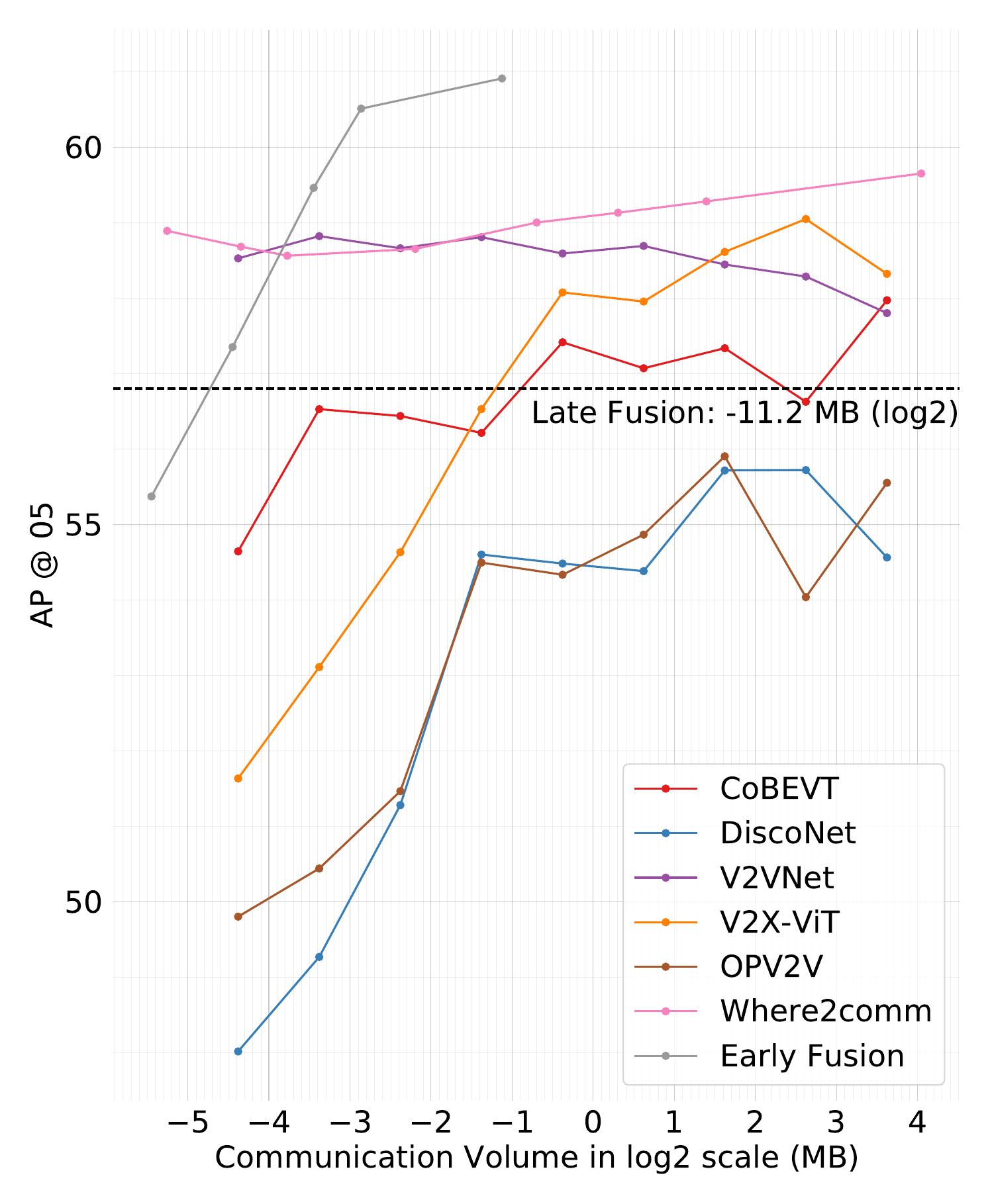}
    }
    \subfloat[V2XSet]{
        \label{fig:compress_v2xset}
        \includegraphics[height=0.295\textheight,trim=12 10 20 10,clip]{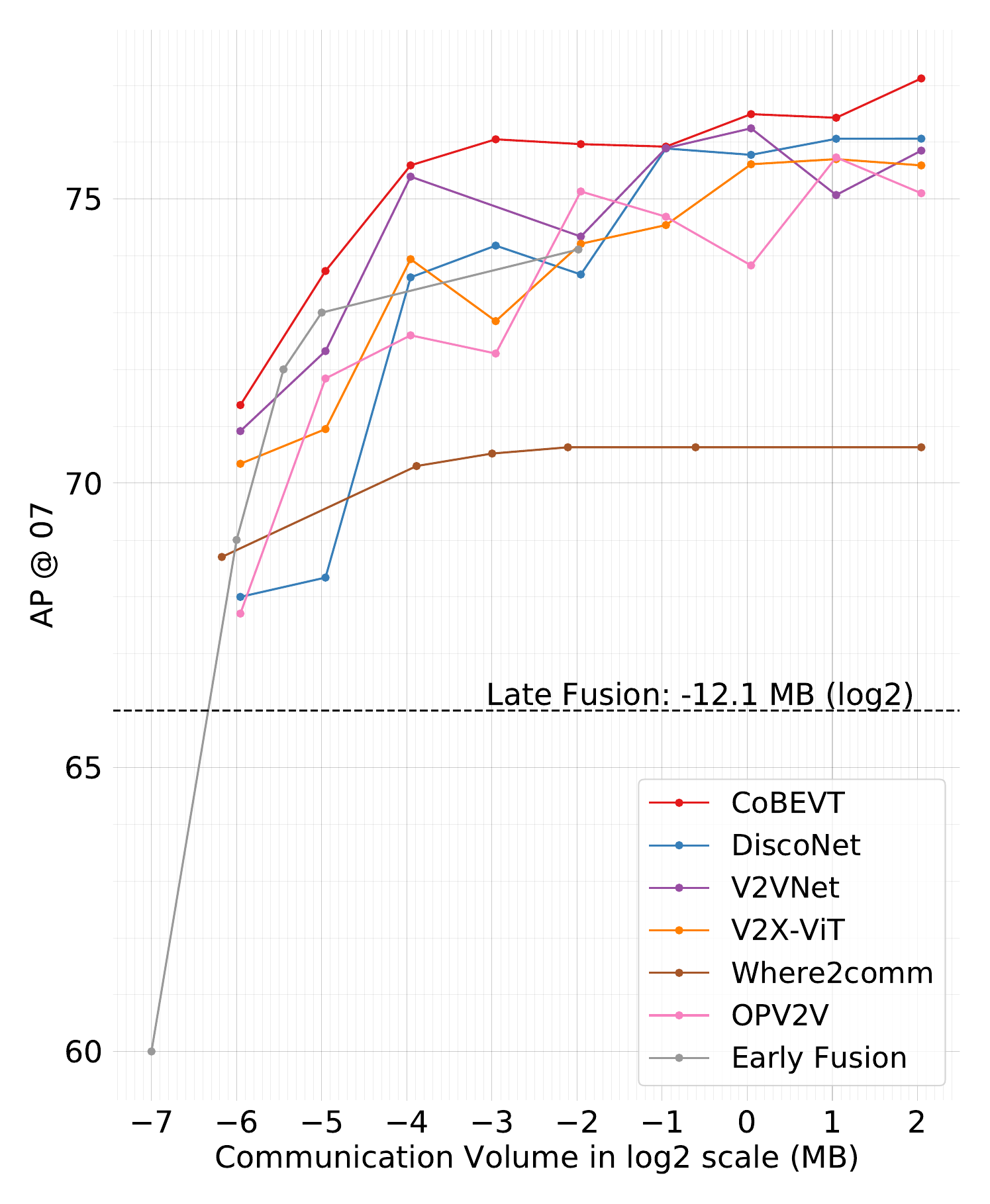} 
    }
    \subfloat[V2V4Real]{
        \label{fig:compress_v2v4real}
        \includegraphics[height=0.295\textheight,trim=12 10 20 10,clip]{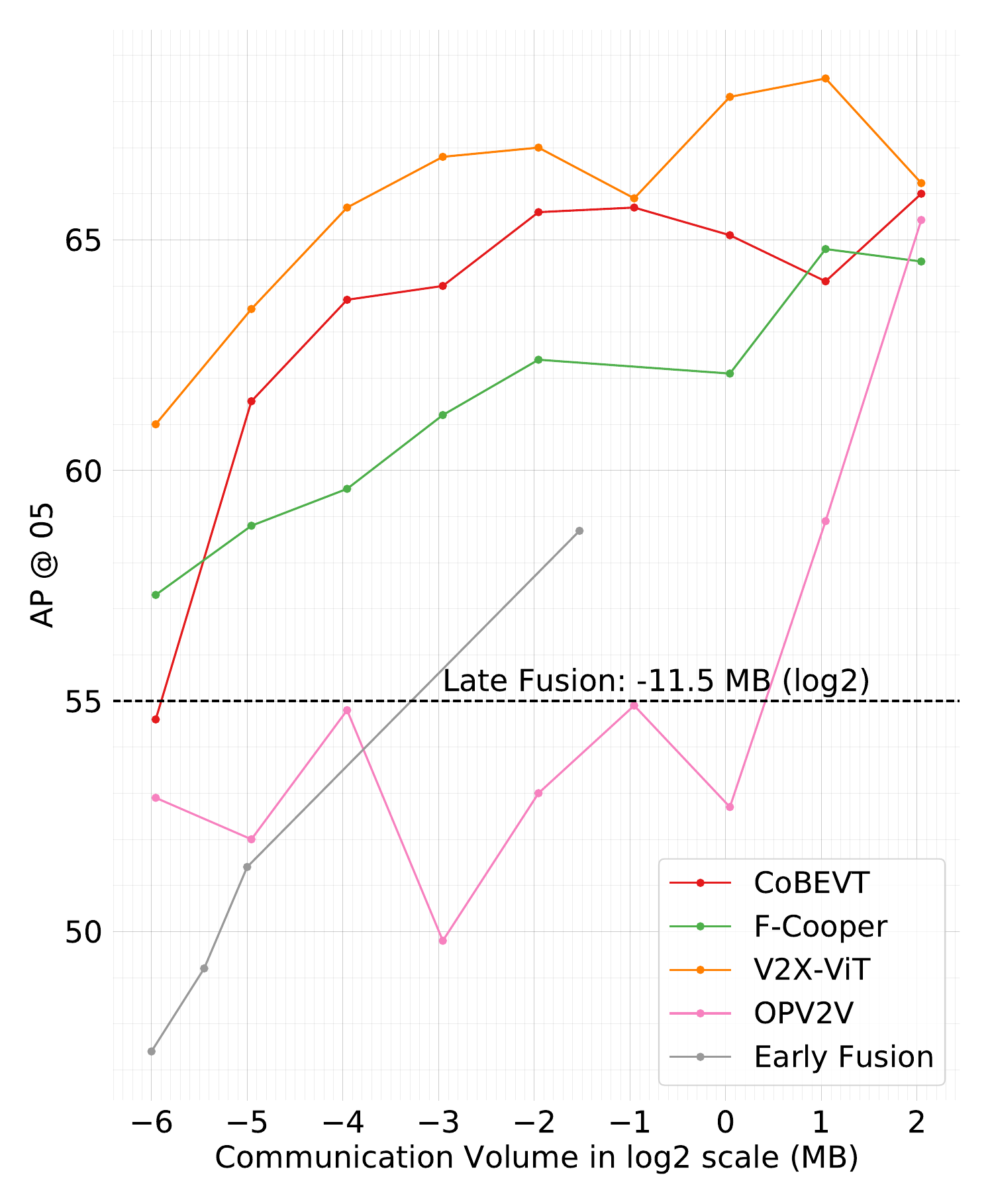} 
    }
    \caption{\textbf{Performance-Bandwidth Trade-off.} The x-axis represents the Bandwidth in log2 scale, and the y-axis represents the AP performance.}
    \vspace{-2mm}
    \label{compress}
\end{figure*}
\subsection{Comparisons of Model Attributes}
\label{subsec:compare}
To compare current CP detection methods, we conduct evaluations on three typical datasets, \ie, OPV2V, V2XSet, and V2X-Sim, presenting in Table~\ref{tab:methods-opv2v},~\ref{tab:methods-v2xset}, and~\ref{tab:methods-v2xsim}, respectively.
For a fair comparison of computational cost, the LiDAR range in OpenCOOD is set to $[-102.4, -38.4, -3.5, 102.4, 38.4, 1.5]$ in all experiments. Similarly, the LiDAR range in CoPerception is set to $[-96.0, -96.0, -3.0, 96.0, 96.0, 2.0]$ in all experiments. For the V2XSet dataset, the \textit{perfect} setting means there is no pose error or communication latency, and the \textit{noisy} setting follows the standard localization error and communication latency in OpenCOOD. 

Table~\ref{tab:methods-opv2v},~\ref{tab:methods-v2xset}, and~\ref{tab:methods-v2xsim} show the model size (Param), computational cost (MACs), transmission size (Transmission), and efficiency (Inference Time) of CP methods on three datasets. Note that \textit{Inference Time} means the computing time of the main network from input to output. 
From these tables, we find that early-fusion and late-fusion methods have lower computational complexity. However, intermediate-fusion methods generally perform better than early-fusion and late-fusion methods, especially under the \textit{noisy} setting.

\subsection{Performance-bandwidth Trade-off Analysis}
\label{subsec:trade-off}
Collaborative perception systems are designed to enhance perception performance by transferring complementary information between different agents. However, in real-world applications, communication bandwidth is typically limited and variable. 
Therefore, to fully utilize the limited and unstable bandwidth, achieving a better balance between transmission volume and precision is crucial for designing CP models. In this section, we explore the performance-bandwidth trade-off of various methods.

For fair comparisons, most intermediate fusion methods such as V2VNet~\cite{wang2020v2vnet}, we leverage $1\times 1$ convolutions to compress and decompress the transmission features along the channel dimension. The channel compression rate is changed from $\times 1 (2^0)$ to $\times 256 (2^8)$, that is, the number of channels of transmission feature maps varying from $256$ to $1$ (The model corresponding to the $\times 2^0$ compression rate does not have a compression module). Besides, Where2comm~\cite{huwhere2comm} adopts a spatial-confidence-aware communication strategy and has a fixed number of channels of $256$.
We first train an uncompressed base model (with $256$ channels) of every method on \textit{perfect} setting. For each compression rate, we fine-tune the model for $3$-$10$ epochs. When training Where2comm, we disable multi-scale and use AttFusion for the fusion module. For early fusion, compression is achieved by Furthest Point Sampling (FPS) without fine-tuning. For point clouds from other agents, only points that fall within the detection range of the ego car are considered, and FPS is applied only to these points. 
All these methods are tested under the \textit{perfect} setting. 
The communication volume is calculated by:
\begin{equation}
\operatorname{Comm}=\log _2(N \times C \times 16 / (8 \times 2^{20}) ), \label{eq1} 
\end{equation} where $N$ represents the number of transmitted elements, $C$ represents the number of channels, $16$ indicates that the data is transmitted in fp16 data type, which incurs almost no performance loss. Then the bit-based volume is converted to megabytes on the $log_2$ scale. In the intermediate fusion methods with channel-based compression, $N$ is typically the full feature map resolution $H\times W$. For early fusion, $N$ is the size of the point cloud to be transmitted and $C$ is $4$ corresponding to the three-dimensional coordinates and the reflectance of the point cloud. For late fusion, $N$ is the number of bounding boxes after NMS, and $C$ is $7$ representing the position of the 3D bounding box.

Fig.~\ref{fig:compress_v2xset} shows the results on V2XSet~\cite{xu2022v2x}. Intermediate fusion methods show no significant performance degradation with the increase of compression ratio. When the compression ratio increased from $\times 2^0$ to $\times 2^5$, most methods suffered less than a $5\%$ decrease in average precision, and the maximum decrease was $4.37\%$ in OPV2V \cite{xu2022opv2v}. These results suggest that an appropriate reduction in the number of channels can effectively eliminate redundancy in collaborative information. Only at high compression ratios ($\times 2^8$, $\times 2^7$, $\times 2^6$), does the performance show a significant decrease, approaching $10\%$.

\begin{table*}
   \caption{\textbf{Robustness evaluation.} 3D object detection results on DAIR-V2X-C, V2XSet, and V2V4Real datasets under various settings, with AP@$0.7$ used for V2XSet and AP@$0.5$ used for DAIR-V2X-C and V2V4Real. Note that `\%' is omitted for simplicity. \textit{Time Delay}: random time delays ranging from $0$ to $500$ms for all agents. \textit{Lossy Communication}: pixel-level replacement of transmitted features using random noise within a fixed range. \textit{Mixed Noise}: combining all aforementioned noises and localization errors (Gaussian noise: mean=$0$, standard deviation=$0.2$).}
   \vspace{-5mm}
   \begin{center}
      \resizebox{\linewidth}{!}{
         \renewcommand\arraystretch{1.3}
    {\fontsize{3.2}{3.5}\selectfont
    \setlength{\tabcolsep}{3.5pt}
    \setlength{\arrayrulewidth}{0.2pt}
    \begin{tabular}{|c|c|c|c|c|c|c|c|c|}
        \hline
        \multirow{2}{*}{Dataset} & \multirow{2}{*}{Methods} & \multicolumn{4}{c|}{Perfect} & \multirow{2}{*}{\shortstack{Time\\[-1.5pt]Delay}} & \multirow{2}{*}{\shortstack{Lossy\\[-2.1pt]Comm.}} & \multirow{2}{*}{\shortstack{Mixed\\[-1.8pt]Noise}}   \\ \cline{3-6}
        & & Overall & 0-30m & 30-50m & 50-100m & &  &   \\
        \hline\hline

        \multirow{9}{*}{DAIR-V2X-C~\cite{yu2022dair}} & CoBEVT~\cite{xu2022cobevt} & 58.0 & 73.4 & 62.5 & 45.8 & 51.9 & 49.1 & 46.8 \\ \cline{2-9}
        & DiscoNet~\cite{mehr2019disconet} & 54.6 & 70.9 & 61.4 & 37.0 & 54.4 & 50.1 & 37.5 \\ \cline{2-9}
        & V2VNet~\cite{wang2020v2vnet} & 57.8 & 73.5 & 62.9 & 45.2 & 50.7 & 57.0 & 49.1 \\ \cline{2-9}
        & V2X-ViT~\cite{xu2022v2x} & 58.3 & 69.7 & 63.9 & 45.1 & 55.1 & 47.4 & 47.8 \\ \cline{2-9}
        & Where2comm~\cite{huwhere2comm} & 59.1 & 73.1 & 65.1 & 48.2 & 54.1 & 47.9 & 45.0 \\ \cline{2-9}
        & OPV2V~\cite{xu2022opv2v} & 55.6 & 70.8 & 60.7 & 43.4 & 48.6 & 40.3 & 37.6 \\ \cline{2-9}
        & Early Fusion & 60.8 & {74.6} & 62.1 & 47.7 & 57.9 & - & - \\ \cline{2-9}
        & Late Fusion & 56.8 & 63.4 & 54.6 & 52.9  & 46.1 & - & - \\ \cline{2-9}
        & No Fusion & 53.9  & 66.5  & 58.2  & 38.9  & - & - & - \\ \hline \hline

        \multirow{9}{*}{V2XSet~\cite{xu2022v2x}} & CoBEVT~\cite{xu2022cobevt} & 77.1 & {93.8} & 81.5 & 56.6 & 52.0 & 66.1 & 49.2 \\ \cline{2-9}
        & DiscoNet~\cite{mehr2019disconet} & 76.1 & 92.6 & 80.8 & 56.4 & 53.1 & 30.3 & 17.5 \\ \cline{2-9}
        & V2VNet~\cite{wang2020v2vnet} & 75.4 & 92.5 & 81.4 & 55.9 & 58.9 & 63.3 & 28.3 \\ \cline{2-9}
        & V2X-ViT~\cite{xu2022v2x} & 75.6 & 92.8 & 78.7 & 53.7 & 60.5 & 50.1 & 48.8 \\ \cline{2-9}
        & Where2comm~\cite{huwhere2comm} & 70.6 & 89.4 & 75.1 & 47.4 & 56.1 & 44.3 & 41.7 \\ \cline{2-9}
        & OPV2V~\cite{xu2022opv2v} & 75.1 & 91.5 & 81.1 & 39.9 & 54.2 & 23.8 & 34.1 \\ \cline{2-9}
        & Early Fusion & 74.1 & 92.0 & 79.3 & 55.4 & 41.4 & - & - \\ \cline{2-9}
        & Late Fusion & 63.4 & 85.2 & 64.6 & 38.7 & 30.2 & - & - \\ \cline{2-9}
        & No Fusion & 44.7  & 71.8  & 48.8  & 16.9                          & - & - & - \\ \hline \hline

        \multirow{6}{*}{V2V4Real~\cite{xu2023v2v4real}} & CoBEVT~\cite{xu2022cobevt} & 66.0 & {81.8} & 53.4 & 45.2 & 55.2 & 56.4 & 48.6 \\ \cline{2-9}
        & F-Cooper~\cite{fcooper} & 64.5 & 79.8 & 51.1 & 45.9 & 54.6 & 5.8 & 5.5 \\ \cline{2-9}
        & V2X-ViT~\cite{xu2022v2x} & 66.2 & 80.3 & {55.2} & {47.3} & 55.2 & 48.0 & 44.8 \\ \cline{2-9}
        & OPV2V~\cite{xu2022opv2v} & 65.4 & 81.2 & 52.0 & 45.6 & 54.5 & 37.9 & 36.2 \\ \cline{2-9}
        & Early Fusion & 58.7 & 76.9 & 40.9 & 46.4 & 49.5 & - & - \\ \cline{2-9}
        & Late Fusion & 55.0 & 73.4 & 43.7 & 36.3  & 47.8 & - & - \\ \cline{2-9}
        & No Fusion   & 41.6  & 65.6  & 29.8  & 8.9         & - & - & - \\ \hline
    \end{tabular}
    }

      }
   \end{center}
   \label{tab:summary}
    \label{tab:range-v2xset}
    \label{tab:range-v2v4real}
   \label{tab:range-dair}
   \label{tab:mix-noise-dair}
   \label{tab:mix-noise-v2xset}
   \label{tab:mix-noise-v2v4real}
   \vspace{-4mm}
\end{table*}
Similar trends can also be observed in DAIR-V2X-C and V2V4Real datasets. As seen in Fig.~\ref{fig:compress_dair} and Fig.~\ref{fig:compress_v2v4real}, on these two real-world datasets, intermediate fusion methods based on channel reduction show only slight performance degradation when the compression ratio is increased from $\times 2^0$ to $\times 2^5$. Only when the compression ratio is increased to $\times 2^7$ and $\times 2^8$ does a significant performance decrease occur.

Where2comm~\cite{huwhere2comm}, which is based on spatial confidence, reduces redundancy in the spatial domain and exhibits smoother performance degradation compared to intermediate fusion methods based on channel compression, as shown in Fig.~\ref{fig:compress_v2xset} and Fig.~\ref{fig:compress_dair}. 
It can be observed that compression based on the sparsity of feature maps in the spatial dimension has finer control granularity compared to compression based on the redundancy in the channel dimension.

For early fusion, the original point cloud data is voluminous and unprocessed, but a significant proportion of the point cloud data is redundant. 
As shown in Fig.~\ref{fig:compress_v2xset} and Fig.~\ref{fig:compress_dair}, on the DAIR-V2X-C and V2XSet datasets, the performance only exhibits a $1\%$ decrease when the point cloud is compressed to about $0.1$ times its original size ($4096$ points). Before this compression ratio, the point cloud still preserves sufficient shape features for the detection algorithm to extract sufficient feature information. As the point cloud is further compressed, the performance rapidly deteriorates, and the curve exhibits a steep linear decline.

Overall, the curves of methods based on spatial confidence are smooth and exhibit potential. The channel dimension also shows significant redundancy and better integration of these two dimensions is a promising direction. Meanwhile, early fusion methods that are not specially designed show competitive performance compared to intermediate fusion methods. In the future, more effective point cloud filtering and compression methods are expected to be applied to early fusion.

\subsection{Comparison in different LiDAR Ranges}
\label{subsec:range}
Analyzing the detection accuracy of targets at different ranges is a common metric for 3D detection tasks. It can reveal the performance change across different distances. For collaborative perception tasks, the backbone is usually the same, and the focus is on analyzing the effects of different fusion modules on detection accuracy. 

We conducted an experiment to investigate the accuracy of object detection results for different target ranges. Specifically, we divided the detection results into three ranges: 0-30m, 30-50m, and 50-100m. The target distance is relative to the ego car. By analyzing the performance of different models across various ranges, we aimed to provide insights into the impact of range on object detection in the context of collaborative perception. It is important to determine whether the fusion modules primarily help to improve the accuracy of nearby occluded targets or to enhance long-range perception accuracy for the ego car.

In Table~\ref{tab:range-dair}, the \textit{Perfect} column presents the accuracy of different methods for different distance targets on the DAIR-V2X-C, V2XSet, and V2V4Real. Comparing the early fusion and the intermediate fusion methods with the no fusion method, both fusion methods show significant improvements in all three ranges. These results demonstrate the fusion modules effectively enhance the detection accuracy for both nearby occlusion and long-range perception scenarios, which is crucial for collaborative perception tasks.

\begin{figure*}[htbp]
    \centering
    \captionsetup[subfloat]{font=scriptsize}
    \subfloat[DAIR-V2X-C]{
        \label{fig:async_dair}
        \includegraphics[height=0.325\textheight,trim=12 10 20 10,clip]{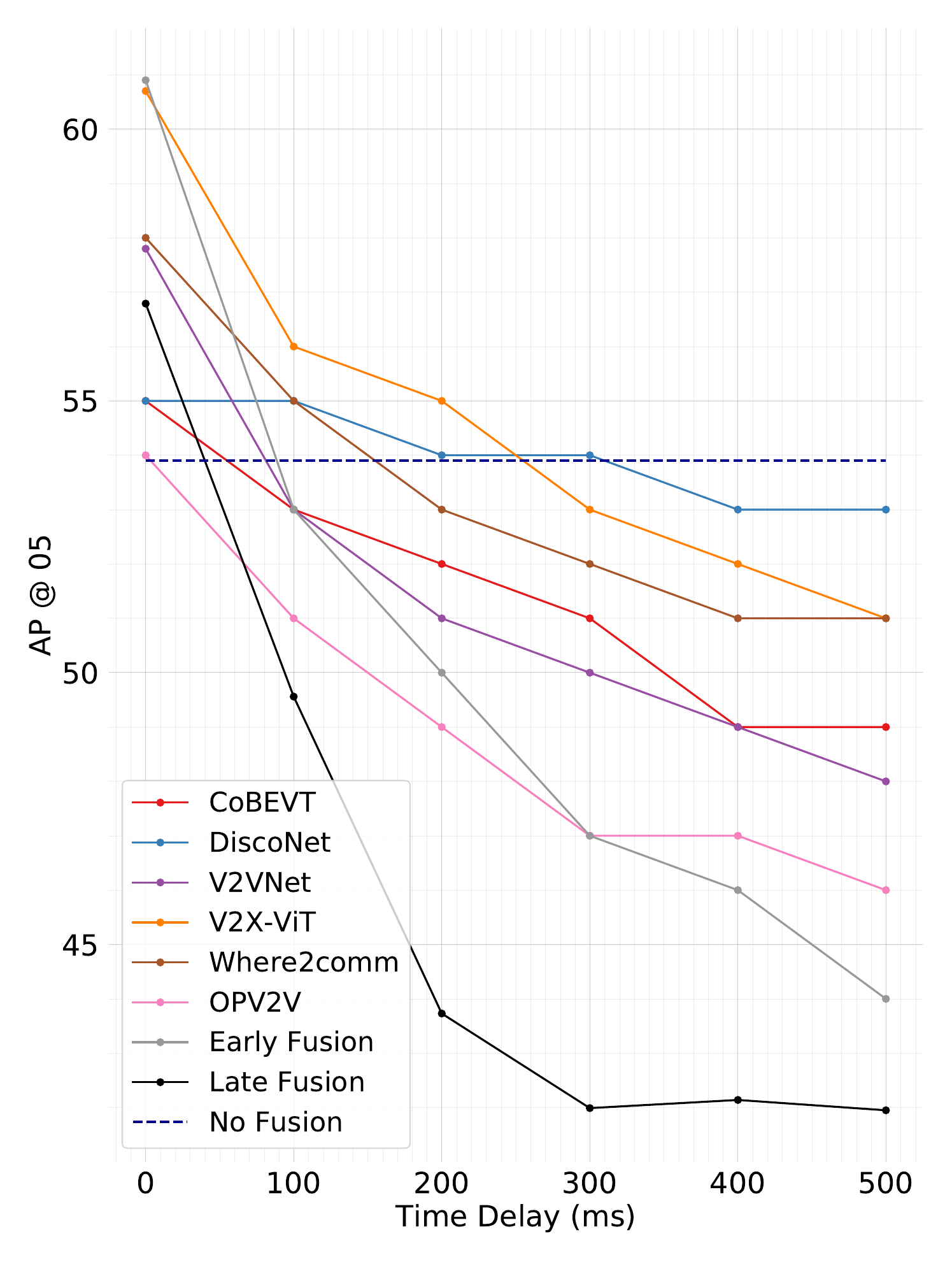}
    }
    \subfloat[V2XSet]{
        \label{fig:async_v2xset}
        \includegraphics[height=0.325\textheight,trim=12 10 20 10,clip]{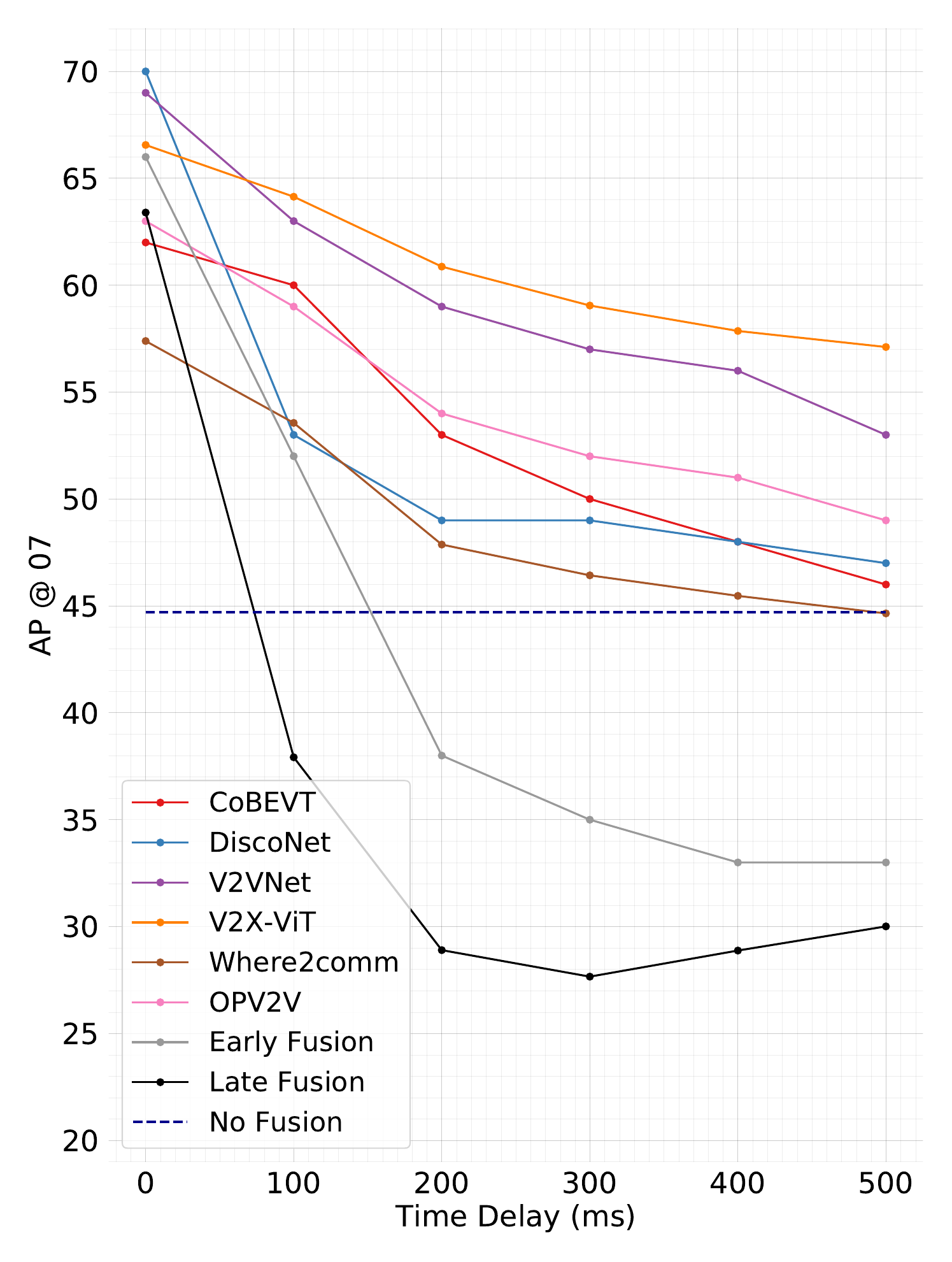} 
    }
    \subfloat[V2V4Real]{
        \label{fig:async_v2v4real}
        \includegraphics[height=0.325\textheight,trim=12 10 20 10,clip]{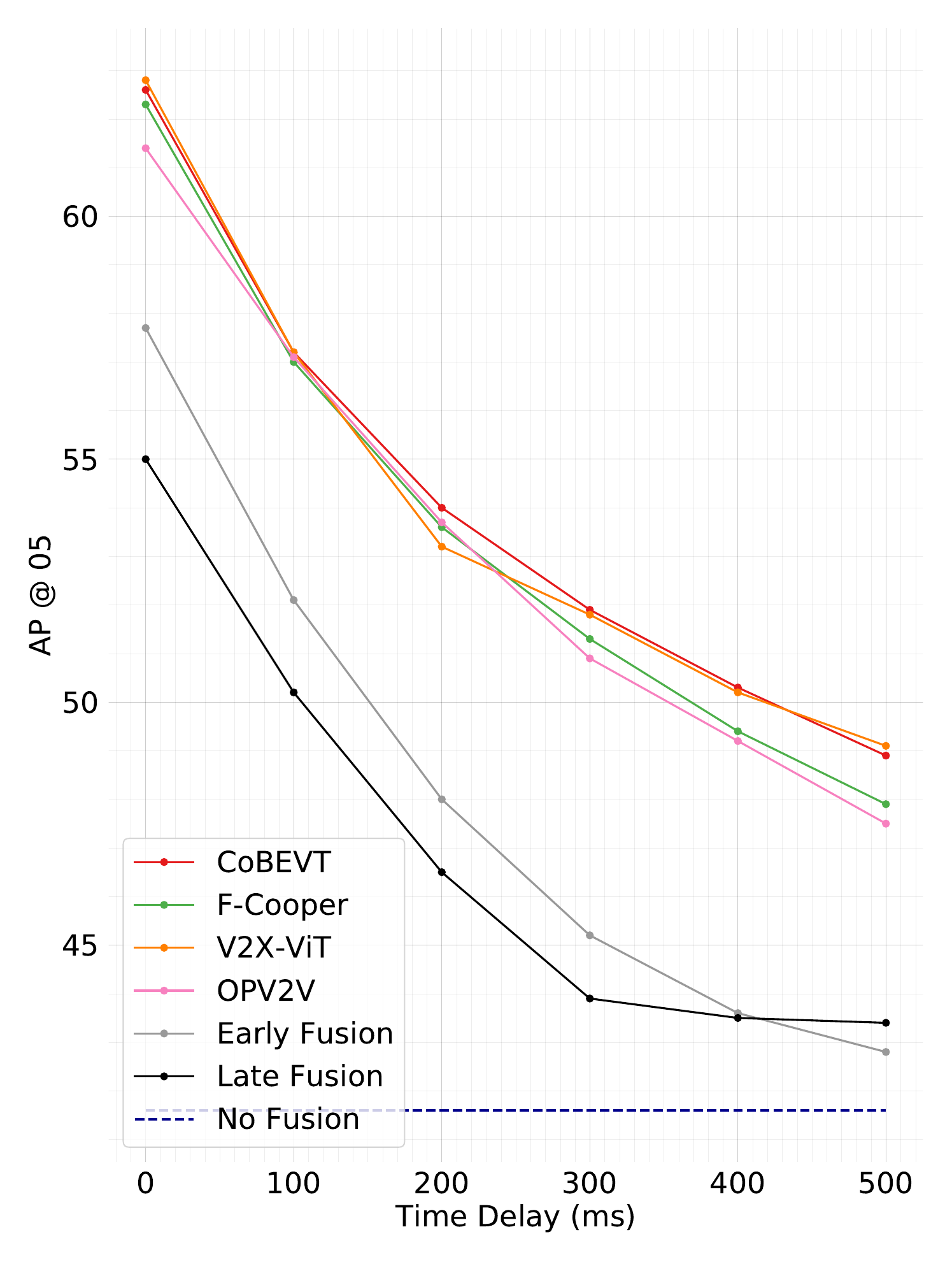} 
    }
    \vspace{-1mm}
    \caption{\textbf{Robustness to communication latency.}
    Performance comparison of models under different time delays. Note that all models are fine-tuned on data with random time delays from $\{100,200,300,400,500\}$.}
    \vspace{-4mm}
    \label{fig:async}
\end{figure*}

\subsection{Robustness Analysis against Time Delay}
\label{subsec:time}
Time delay, or communication latency, is a pervasive problem in real-world V2X communication. It induces an asynchronization between ego features and received collaborative features. Practical collaborative perception methods are expected to be robust against time delays.

In this section, we explore the model robustness against time delay.
We fine-tuned methods based on the model trained in \textit{perfect} setting, with latency uniformly sampled from $0$ to $500$ms. During testing, we adopt fixed and random latency scenarios. In fixed-delay scenarios (Fig.~\ref{fig:async}), we set a fixed delay (ranging from $0$ to $500$ms) for each agent to transmit its features to the ego car. In random-delay scenarios (Table~\ref{tab:summary}), we set delays sampled from a uniform distribution between $0$ and $500$ for each agent.

Under the aforementioned conditions, the performance of all models decreases accordingly. As shown in Fig.~\ref{fig:async_v2xset} and the \textit{Time Delay} column in Table~\ref{tab:mix-noise-v2xset}, V2X-ViT \cite{xu2022v2x} exhibit strong robustness to latency in V2XSet. Specifically, in random-delay scenarios, the performance degradation is only $19.9\%$, while others drop more than $20\%$. Furthermore, in fixed-delay scenarios, the performance degradation of V2X-ViT is less than $4\%$ for every additional $100$ms of latency. In contrast, compared with intermediate fusion methods, early fusion and late fusion exhibit more severe performance degradation, indicating poor robustness to latency. When the time delay increases to $200$ms, APs of early fusion and late fusion decrease by $48.64\%$ and $54.4\%$, respectively. 

The results on DAIR-V2X-C are presented in Fig. \ref{fig:async_dair}, DiscoNet demonstrates the best robustness and performance degradation of only $0.37\%$ under random latency scenarios. However, the conclusions on V2V4Real (Fig. \ref{fig:async_v2v4real}) are different, as all three kinds of fusion methods show a relatively severe performance degradation with similar magnitudes.

Overall, the results indicate that latency is a critical factor that affects the performance in V2X perception. Moreover, the results highlight the importance of designing fusion methods that are robust to transmission latency to ensure reliable and accurate perception in real-world scenarios.

\begin{figure*}[htbp]
    \centering
    \captionsetup[subfloat]{font=scriptsize}
    \subfloat[DAIR-V2X-C]{
        \label{positional_error_dair}
        \label{heading_error_dair}
        \label{error_dair}
        \includegraphics[height=0.24\textheight,trim=12 5 5 10,clip]{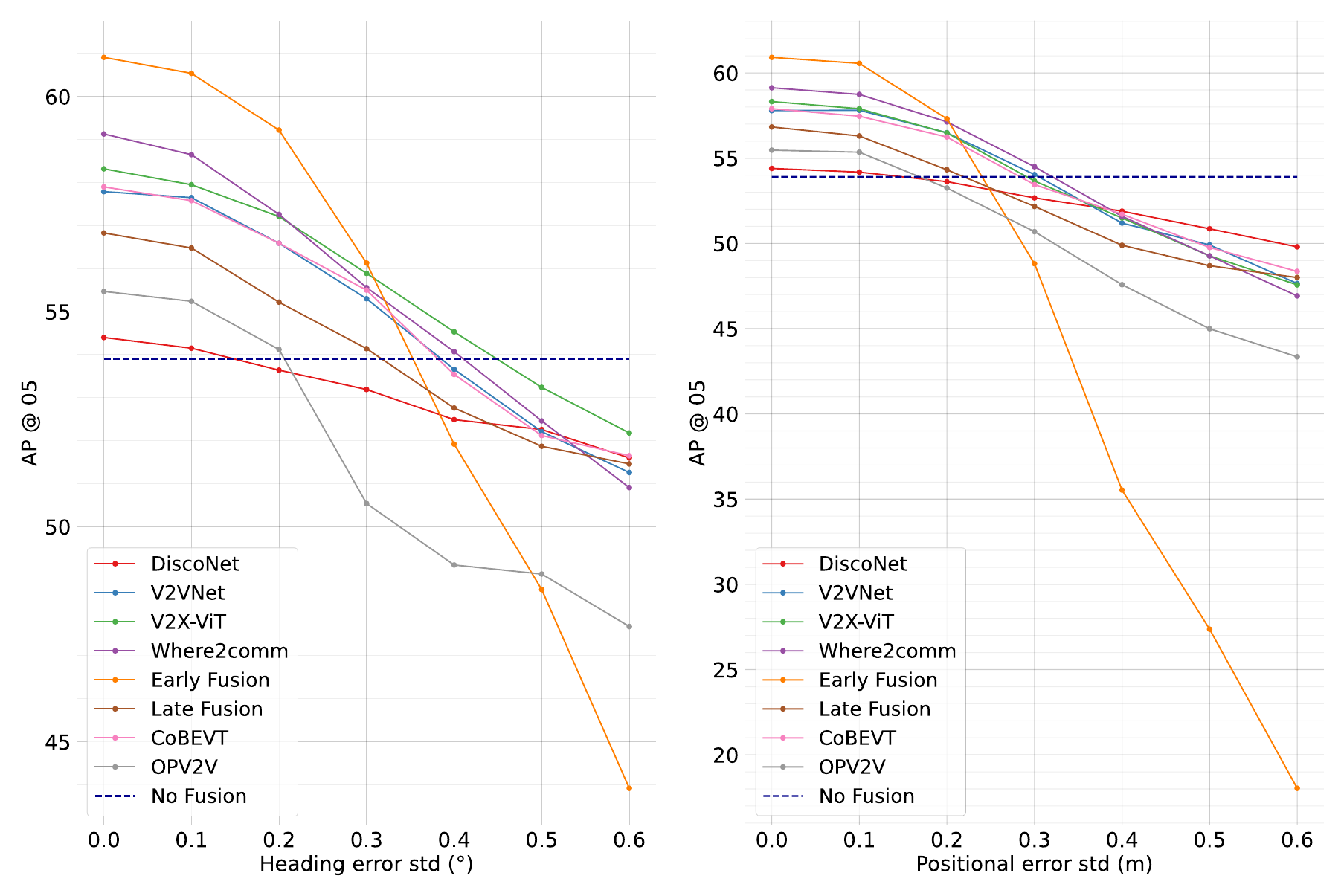}
    }
    \subfloat[V2XSet]{
        \label{positional_error_v2xset}
        \label{heading_error_v2xset}
        \label{error_v2xset}
        \includegraphics[height=0.24\textheight,trim=12 5 5 10,clip]{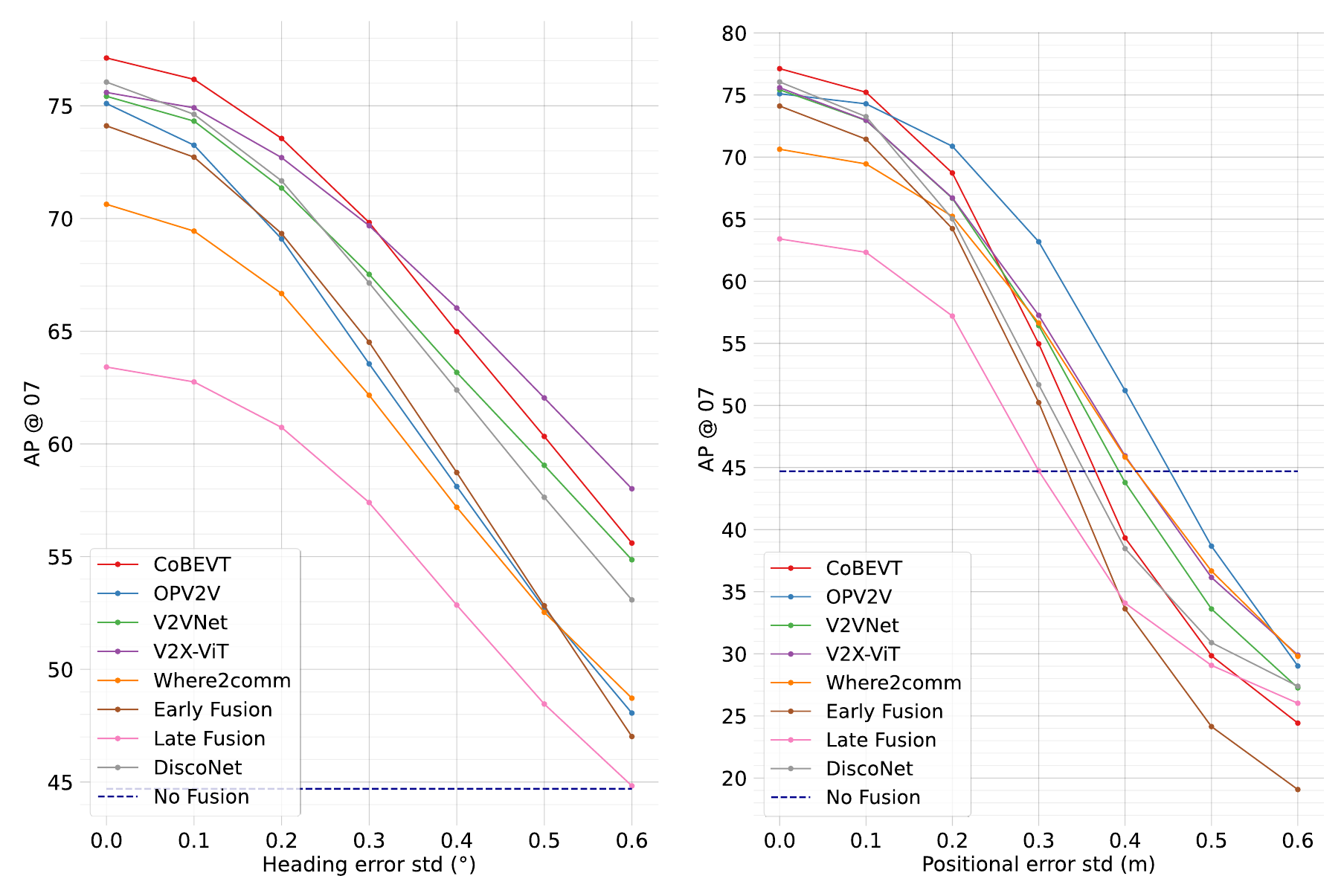} 
    }
    \caption{\textbf{Robustness to localization errors.} Models are trained in a perfect setting and directly apply to localization errors setting, \ie, Gaussian noise with a mean of $0$ applied to the LiDAR pose. The noise levels are gradually adjusted by varying the standard deviation for both heading error (RYP dimensions) and positional error (XYZ dimensions), respectively.}
    \vspace{-4mm}
    \label{fig:positional_error}
\end{figure*}

\subsection{Robustness Analysis against Localization Error}
\label{subsec:pose-error}
Cooperative perception models rely on the relative coordination between two collaborative agents to fuse their information. Consequently, the quality of localization essentially affects the process of information fusion. Though advanced localization technologies (\eg, GPS) have been widely facilitated in perception systems, localization error is inevitable. In this way, it is critical to achieve cooperative perception models that are robust to localization error. 

To facilitate relative research, we conduct experiments on two datasets (\ie, DAIR-V2X-C \& V2XSet), where cooperative perception models are trained in the environment of perfect localization while tested in an environment with simulated localization noises without any fine-tuning. The noises are sampled from Gauss Distribution, with a mean of $0$ and a changeable standard deviation. 

Experimental results based on DAIR-V2X-C are exhibited in Fig.~\ref{error_dair}. Compared to intermediate fusion and late fusion methods, early fusion is much more sensitive to localization error, whose AP@$0.5$ drops $70.5\%$ when position error std is $0.6$m. Besides, though weaker than other intermediate fusion methods when no localization error exists, Disconet has minor performance degradation when the error worsens. Such phenomenon indicates that when elaborately designed fusion methods improve performance in perfect settings, robustness to localization error may hurt.

Fig.~\ref{error_v2xset} shows the results on V2XSet. Generally, heading error has less impact on the performance compared to positional error. Performance on early fusion and late fusion methods drop violently with localization noises increasing, where intermediate fusion methods show better robustness.

\subsection{Robustness Analysis against Lossy Communication}
Most collaborative perception methods assume an ideal communication and do not consider the impact of lossy communication on feature sharing. Lossy communication such as packet loss and signal interference is common in the real world~\cite{wang2021digital,zeadally2020tutorial,yan2022discrete,sun2021survey}, leading to inaccurately shared intermediate features and thus harming the collaborative perception performance.

In this section, we investigate the impact of lossy communication on different collaborative detection methods. Initially, We train the model under ideal communication without any compression (\ie, in \textit{perfect} setting) and then finetune it in a lossy communication environment. To simulate the impact of lossy communication on collaborative perception, we adopt the approach proposed by Li et al. \cite{li2023learning}, which uniformly samples a probability $p$ for the intermediate features used in collaboration. Elements in the intermediate features within a fixed noise range $[0, 29.5]$ (determined by the original intermediate feature range) are then replaced with random noise with the probability of $p$.

As seen in the \textit{Lossy Comm.} column in Table~\ref{tab:mix-noise-dair}, lossy communication has a negative impact on most models, particularly on the simulated dataset V2XSet. 
Where2comm gets $44.3\%$ in V2XSet under lossy communication, a $37.2\%$ decrease, and $47.9\%$ in DAIR-V2X-C, an $18.9\%$ decrease. 
V2X-ViT gets $50.1\%$ in V2XSet, a $33.7\%$ decrease, and $47.4\%$ in DAIR-V2X-C, a $18.7\%$ decrease. 
Some models exhibited higher robustness. V2VNet gets $63.3\%$ in V2XSet, a $16.04\%$ decrease, and $57.0\%$ in DAIR-V2X-C, a $1.4\%$ decrease. CoBEVT achieves $66.1\%$ in V2XSet, a $14.3\%$ decrease.

To sum up, most CP methods are generally susceptible to lossy communication. Given that imperfect communication is a real presence in the real world, it is of great significance to devise methods that are robust to lossy communication.

\subsection{Robustness Analysis against Mixed Noises}
\label{subsec:mix}
In order to simulate real-world scenarios more realistically, we consider a mixed noise setting that combines time delay, localization error, and lossy communication to evaluate how various models perform under these combined noises. Similarly, we fine-tune well-trained models without any compression under the mixed noise setting, keeping the settings for time delay, positional noise, and lossy communication consistent with those in the previous sections.

As shown in the \textit{Mixed Noise} column in Table~\ref{tab:mix-noise-dair}, CoBEVT~\cite{xu2022cobevt}, DiscoNet~\cite{mehr2019disconet}, V2VNet~\cite{wang2020v2vnet}, V2X-ViT~\cite{xu2022v2x}, Where2comm~\cite{huwhere2comm} and OPV2V~\cite{xu2022opv2v} exhibit varying degrees of performance decrease in both DAIR-V2X-C~\cite{yu2022dair} and V2XSet~\cite{xu2022v2x}.
In general, in realistic scenarios that involve various types of noise, current collaborative perception models exhibit a significant decrease in accuracy. Effectively handling complex noise remains a challenging task.
\begin{table}
   \caption{\textbf{Generalization evaluation.} \textcolor{mygray}{Gray texts} refer to the AP drop when the model is trained on V2XSet~\cite{xu2022v2x} but evaluated on DAIR-V2X-C~\cite{yu2022dair}.}
   \vspace{-3mm}
   \begin{center}
      \resizebox{0.45\textwidth}{!}{
         \renewcommand\arraystretch{1.3}
    \footnotesize
    \setlength{\tabcolsep}{3.5pt}{
        \begin{tabular}{ccccc}
		\hline
            Method & Gap & AP @ 0.3 & AP @ 0.5 & AP @ 0.7\\
		\hline
            \multirow{2}{*}{F-Cooper~\cite{fcooper}} & \ding{56} & 61.26 & 57.19 & 45.04\\
            & \ding{51} & 35.12 \textcolor{mygray}{(-26.14)} & 30.27 \textcolor{mygray}{(-26.92)} & 13.46 \textcolor{mygray}{(-31.58)}\\
            \arrayrulecolor{mygray}\hline
            \multirow{2}{*}{CoBEVT~\cite{xu2022cobevt}} & \ding{56} & 56.31 & 52.11 & 38.96\\
            & \ding{51} & 28.09 \textcolor{mygray}{(-28.22)} & 25.05 \textcolor{mygray}{(-27.06)} & 11.60 \textcolor{mygray}{(-27.36)}\\
            \hline
            \multirow{2}{*}{AttFusion~\cite{xu2022opv2v}} & \ding{56} & 60.86 & 56.29 & 44.23\\
            & \ding{51} & 33.49 \textcolor{mygray}{(-27.37)} & 29.80 \textcolor{mygray}{(-26.49)} & 15.54 \textcolor{mygray}{(-28.69)}\\
            \hline
			\multirow{2}{*}{V2X-ViT~\cite{xu2022v2x}} & \ding{56} & 62.69 & 59.02 & 45.61\\
            & \ding{51} & 34.73 \textcolor{mygray}{(-27.96)} & 31.23 \textcolor{mygray}{(-27.79)} & 14.94 \textcolor{mygray}{(-30.67)}\\
            \hline
            \multirow{2}{*}{Where2comm~\cite{huwhere2comm}} & \ding{56} & 58.52 & 53.58 & 36.95\\
            & \ding{51} & 31.59 \textcolor{mygray}{(-26.93)} & 25.66 \textcolor{mygray}{(-27.92)} & 9.06 \textcolor{mygray}{(-27.89)}\\
		\arrayrulecolor{black}\hline
	\end{tabular}
    }
      }
   \end{center}
   \vspace{-4mm}
   \label{tab:da}
\end{table}
\subsection{Sim-to-Real Generalization Evaluation}
\label{subsec:sim2real}
Models are supposed to be robust across different scenarios, \eg, geographical location, number of collaborative agents, and sensor type and placement.
However, acquiring training data under multiple settings is not only time-consuming and expensive but also hard to cover all possible cases in real applications. 
Consequently, the generalization ability of collaborative perception models is especially critical.

In this section, we assess the generalization ability of several models by training on a simulated dataset V2XSet and evaluating on a real-world dataset DAIR-V2X-C with complemented annotations from CoAlign~\cite{lu2023robust}. The two datasets differ in LiDAR type and placement, number of collaborative agents, \etc, resulting in a significant domain gap. 
For comparison, we directly train these models on DAIR-V2X-C and evaluate them on the same dataset without the domain gap. We assume a \textit{perfect} setting without time delay and localization error in this experiment. In addition, we only consider labels with \textit{Car} type in both training and evaluation. As Table~\ref{tab:da} reveals, all selected methods have a significant precision drop (around $30\%$) when there is a domain gap between training and inference. This performance degradation calls for models with stronger generalization ability and effective domain adaptation/generalization methods in collaborative perception.

\begin{figure}
   \begin{center}
      \includegraphics[width=1\linewidth]{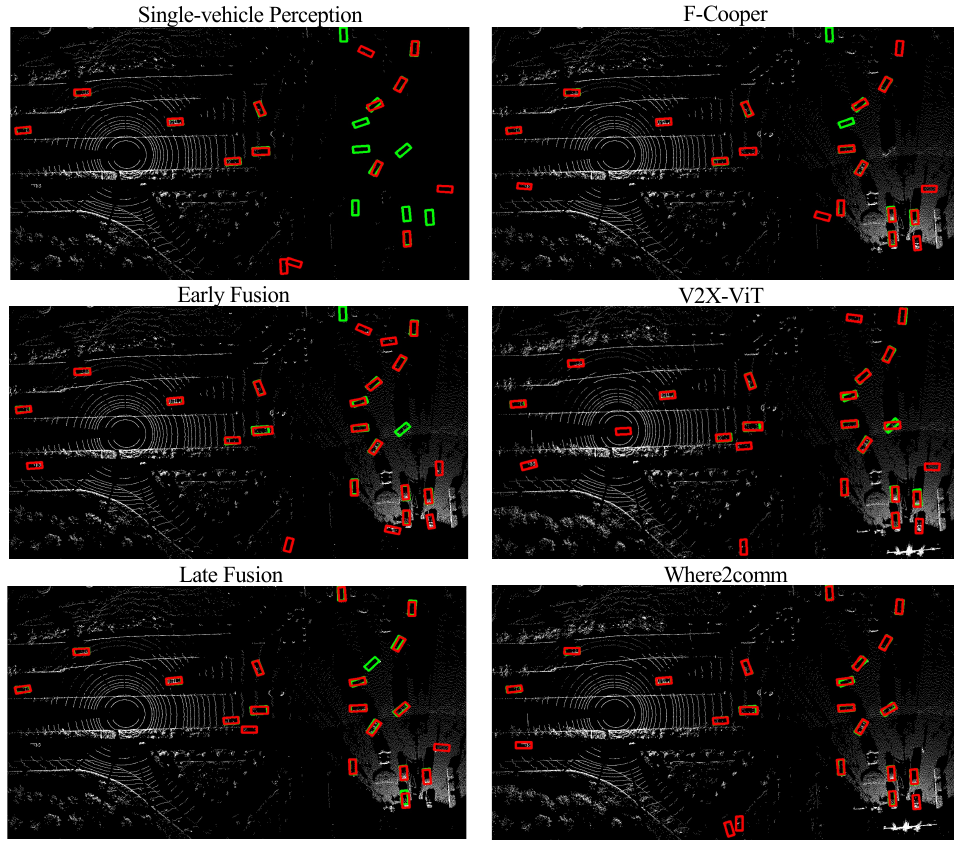}
   \end{center}
   \vspace{-5mm}
      \caption{Visualizations on DAIR-V2X-C.}
      \vspace{-3mm}
   \label{fig:vis_dair}
\end{figure}
\begin{figure}
   \begin{center}
      \includegraphics[width=1\linewidth]{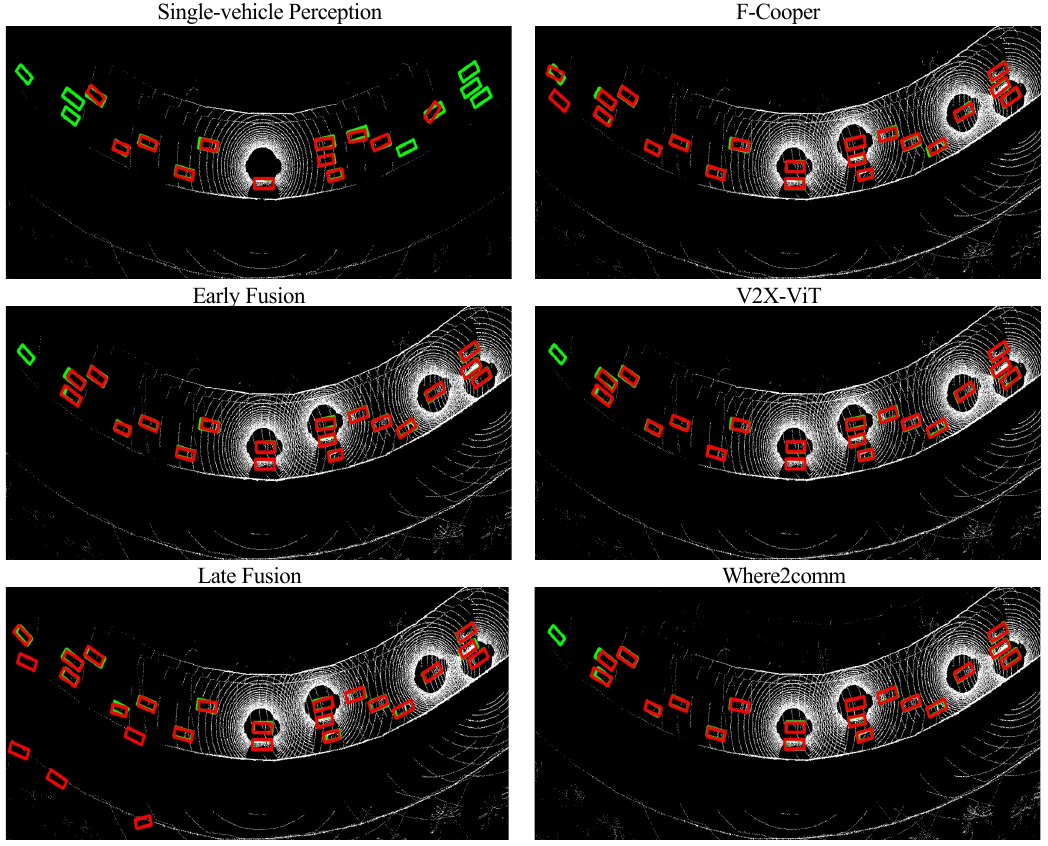}
   \end{center}
   \vspace{-5mm}
      \caption{Visualizations on V2XSet.}
      \vspace{-3mm}
   \label{fig:vis_v2xset}
\end{figure}

\subsection{Qualitative Analysis}
\label{subsec:qualitative}
To analyze cooperative perception methods in a more intuitive way, we visualize certain detection results on DAIR-V2X-C (Fig.~\ref{fig:vis_dair}), V2XSet (Fig.~\ref{fig:vis_v2xset}). In each image, we use green boxes to denote ground truths and red boxes for predictions.

Fig.~\ref{fig:vis_dair} shows a scene in DAIR-V2X-C, where the ego-car is approaching the intersection. Basically, this scene includes two LiDAR sensors, one from the ego-car on the middle road and the other from roadside infrastructure at the intersection. Without information from the roadside infrastructure, No Collaboration method fails to detect cars at the intersection due to long distances and occlusions. Compared to other methods, V2X-ViT~\cite{xu2022v2x} and Where2comm~\cite{huwhere2comm} outputs more accurate detection results, yet the predictions of the heading angle need further improvement.  

Fig.~\ref{fig:vis_v2xset} presents the detection results on a curve from V2XSet. There are four collaborative cars in this scene, mainly distributed in the right region. Apart from the No Collaboration method, all other methods detect the cars on the right bottom successfully, showing strengths of cooperative perception. In contrast, detection performance is poorer on the left region, since no collaborative car exists here. Besides, compared to early fusion and intermediate fusion, late fusion tends to give out more false alarms.

\section{Challenges and Future Work}
\label{sec:future}
In this section, we present an in-depth discussion about the remaining challenges and issues. Accordingly, we discuss some promising research directions that may help to further advance the development of V2X collaborative perception. 
\subsection{Performance-bandwidth Trade-off}
As mentioned in \S\ref{subsec:trade-off}, the trade-off between bandwidth and performance is crucial for CP methods in V2X scenarios. 
Typically, methods for reducing bandwidth can be categorized into two modes, \ie, \emph{selection-based} and \emph{compression-based}, aiming to eliminate spatial and channel redundancies. Currently, there are few methods that consider spatial redundancy, with most methods focusing only on channel redundancy by designing fusion modules and auto-encoders. Therefore, a promising direction is to simultaneously consider reducing spatial and channel redundancies.

Besides, for each compression rate, it needs to conduct model fine-tuning, which is time-consuming and results in coarse compression rate control. 
Furthermore, in the field of learning-based data compression, methods like variable bit rate, bit allocation (feature flow), thumbnail, and distributed coding can be adopted and modified for V2X CP models. These methods can further refine the allocation of compression rates by assigning different compression rates for different data in the spatial and temporal dimensions according to their importance. Also, adapting appropriate compression rates based on agents' transmission environments is another direction.
Thus, future works may focus on exploring to improve the allocation of compression rates and improve the compression efficiency and adaptability of collaborative perception systems.


\subsection{Simulator}
The physical asset models (\eg, vehicles, buildings) of current simulators generally have a large gap with the real environment, and the sensor models in the simulators can hardly represent some common phenomena in the real world, such as camera dynamic blur and motion distortion of LiDAR. Also, most simulators ignore the noise effects of different weather on sensor data and data transmission process.
In the future, the simulator can also be explored in high-realism assets, communication transmission, and scene editing to reduce the domain gap of sensor data and diversify scenarios. A feasible way is to use AIGC techniques to enhance the simulator, such as using NeRF to generate 3D assets, building a more realistic library of physical assets, and rendering simulation data secondarily through a domain migration approach based on generation techniques. There is already some work to simulate LiDAR noise in different weather~\cite{hahner2022lidar} and a simulation platform to reconstruct realistic scenarios using NeRF~\cite{yang2023unisim}.

\subsection{LiDARs Placement}
To optimize the LiDAR perception performance, the placement of LiDAR sensors is critical~\cite{hu2022investigating}.
Recently, some literature has considered the LiDAR perception problem from the perspective of LiDAR placement~\cite{hu2022investigating,ma2021perception}, which is a new perspective and crucial, since improper LiDAR placements may cause poor quality sensing data and thus lead to inferior perception performance.
With the rapid development of V2X applications, it is necessary to know how to choose the optimal placement to maximize the benefits of infrastructure sensors.
Existing research on LiDAR placement mainly focuses on vehicles, in the future, the placement problem for infrastructure sensors is eager to be considered. Since the placement of infrastructure, LiDAR has a higher degree of freedom, where the location positions and roll, pitch, and yaw angles need to be considered.
On the other hand, existing literature proposes methods to evaluate different LiDAR placements instead of directly optimizing and searching sensor positions. 
A feasible future research direction will be to directly optimize the number of LiDARs and the placement of each LiDAR to achieve superior perception performance.

\subsection{Datasets and Collaboration Graph}
Current CP datasets tend to have about $5$ collaborating agents as mentioned in \S\ref{sec:datasets}, leading to a \emph{sparse distribution} of agents across a vast spatial range. However, the prevalence of connected vehicles in future V2X scenarios is expected to increase significantly.
Consequently, a promising direction is to develop datasets that feature a larger number of collaborating agents within a single frame, thereby achieving a \emph{dense distribution} of agents on the road. This direction would align better with future scenarios and enable researchers to explore more comprehensive solutions.

Most current works assume building a fully connected graph, where the ego car receives information from all agents. This is acceptable with a limited number of CAVs (about 5) in existing CP datasets. However, when the number of CAVs increases significantly, the resource consumption of a dense collaboration graph becomes unacceptable. 
In this case, the edges of the collaboration graph should be sparse rather than a fully connected graph. Therefore, future works can consider how to build the sparse graph and the transmitted features based on the entire scene. Agents may need to exchange small meta-information to aid decision-making or elect a super-node to synthesize various information. 
Additionally, a dataset with dense CAV scenarios may advance research, and sparse CAV scenarios can be randomly sampled from dense scenes. 
Future work can also explore advanced graph learning techniques to improve the performance of CP in dense graph scenarios. 

\subsection{Transmission Scheme}
Currently, the information transmission scheme of V2X is not efficient enough. When each agent transmits data, it either broadcasts or unicasts multiple copies of the data to all other agents, which results in high bandwidth overhead and high communication latency. Therefore, a potential solution is to use the roadside unit as a routing node to collect and distribute information uniformly, which can effectively reduce bandwidth and latency.

In addition, relying on roadside units as the important routing node may limit flexibility, \eg, some roads do not contain intelligent roadside units. Therefore, a better way is to dynamically select a CAV from the V2X network as a logical routing node based on factors such as distance, computability, bandwidth \etc~ In this way, the vehicle that serves as movable logical routing nodes can perform information collecting and distributing instead of fixed roadside units.
Besides, methods relying on the routing node based transmission paradigm may face the potential issue of cross-region handoff, \ie, when a vehicle switches between two areas managed by different routing nodes, which can also be regarded as a valuable research direction.

\subsection{Generalization}
\subsubsection{Sim2Real}
Training CP models require a large amount of annotated data. However, collecting real-world datasets is time-consuming and expensive, and current datasets typically only involve a small number of CAVs or infrastructures for collaborating, making it difficult to meet various requirements in real scenarios, such as dense CAV scenarios and adverse weather conditions. 
A suitable approach is to use simulation tools such as CARLA~\cite{dosovitskiy2017carla} to generate simulated collaborative perception data.
However, simulation data and real-world data differ in many aspects such as sensor types, reflection patterns, and road environments. Significant sim2real domain gaps result in differences in point cloud density, distribution, reflectivity, \etc~posing a challenge to the model's generalization ability. 
Current works have made preliminary experiments in sim2real for V2V communication~\cite{xu2023v2v4real}. Future work can focus on improving the simulation fidelity and diversity, as well as enhancing the generalization ability of existing methods. 

\subsubsection{Towards Practical Cooperative Perception}
CP system in real-world scenarios faces complex noise challenges, including time delays, positional noise, lossy communication (packet loss, communication noise and interruption), \etc~Regarding time delays, although encoding time delays during training can enable the model to perceive and correct features without significant additional computation, combining temporal information and time delay compensation may be more conducive to predicting accurate features. 
SyncNet~\cite{lei2022latency} made pioneer attempts to leverage historical information to compensate for time delays, but inferring each frame requires extracting all historical information, which may significantly increase inference time. 
Therefore, using a sliding window-like method to dynamically handle temporal information over time may be a better choice, maintaining a global feature to memorize the historical information outside the window. 
Besides, integrating the temporal detection module into the pipeline of time delay compensation may be another promising direction.

In addition, current large-scale datasets lack sufficient diversity and complexity, where sensors in the real world may generate low-quality data due to various noises such as adverse weather. To build a more robust system, it is urgent to collect data in complex environments and propose methods accordingly. 
Current work~\cite{xiang2022v2xp} has studied adversarial data generation in simulated scenarios. Future works can also construct more realistic scenarios and fill the gap between simulation and reality further.

Another promising direction is to explore multi-domain continuous learning techniques to improve the generalization ability in unseen environments. Multi-domain continuous learning techniques can enable the perception model to learn and adapt from new domains over time while retaining the knowledge learned from previous domains.

\subsection{Fusion Methods}
\subsubsection{Early Fusion}
Early fusion has been challenged for its high bandwidth requirements. However, for perception tasks, a high density of point clouds is not always necessary. Experiments in \S\ref{subsec:trade-off} show that simply downsampling the raw point cloud can achieve a competitive bandwidth-accuracy trade-off compared to intermediate fusion. 
Inspired by this observation, future works can modify the advanced point cloud compression methods for early fusion methods, further improving the performance while reducing bandwidth. 
In addition, enhancing the robustness of the raw point cloud to various real-world noises through end-to-end optimization or deep point cloud compression is also an interesting direction.

\subsubsection{Intermediate Fusion}
One major challenge hindering the practicality of intermediate fusion is how to achieve collaboration between different intermediate fusion models. It is unrealistic for all vehicles to deploy the same model. Even for the same company, different models may exist due to different software versions on the vehicles. When shared features come from different models, there is a significant domain gap, which can easily lead to performance degradation. Although preliminary exploration has been conducted in this regard, there is still considerable room for improvement in the performance of cross-model collaboration and system design.

\subsubsection{Late Fusion}
Instead of directly transmitting object boxes, future late fusion methods can transmit the soft probability distribution of objects, which may further improve the performance upper bound with only a minimal increase in bandwidth usage. 
Ego-car can use the probability distribution to perform anchor initializing, boxes refining, \etc~ 
However, there is still room for improvement in terms of the robustness and generalization performance of late fusion in challenging environments such as significant localization errors and communication noises. 
An interesting research direction is to combine the advantages of the three fusion schemes to achieve a multi-stage fusion architecture. 

\subsection{Security and Privacy}
In order to participate in the V2X system, a car inevitably shares information with other cars and infrastructures. 
Such connections can lead to several hidden dangers: (\romannumeral1) Privacy of users may leak through the information transmissions; (\romannumeral2) Hackers may invade the V2X network and perform attacks on the system, which may result in deliberate traffic accidents. 
Therefore, the V2X system should be strictly protected to prevent the above situations.

Some strategies may contribute to the solution. For example, Privacy-preserving computing techniques can be applied in CP methods to protect privacy within the data. Besides, data can be encrypted before transmissions to ensure privacy and safety, and adversarial attack and defense techniques can be applied in V2X systems. Additionally, an independent emergency driving plan should be designed for autonomous driving cars, which is enabled when the V2X system is hacked. Also, related laws should be set to promote the applications of V2X autonomous driving. 

\section{Conclusion}
\label{sec:conclusion}
In this survey, we present a comprehensive review of collaborative perception toward vehicle-to-everything autonomous driving scenarios. 
We begin with a brief introduction to V2X autonomous driving and collaborative perception. 
To provide a systematic view of V2X autonomous driving, we present a typical architecture and workflow of V2X systems and describe the role of collaborative perception within the whole system.
Besides, we summarize and compare existing datasets for collaborative perception. 
For taxonomy, we categorize collaborative perception methods from various perspectives. 
Moreover, we perform extensive experiments to evaluate existing methods including the model efficiency, robustness, generalization, \etc,~and thoroughly discuss the results to provide unexplored insights. 
Finally, we look into the open challenges and issues, offering potential directions and solutions for future research. 


%





\ifCLASSOPTIONcaptionsoff
  \newpage
\fi



\bibliographystyle{IEEEtran}
\bibliography{trans}
%

%



\vfill


\end{document}